\documentclass{article} 
\usepackage{iclr2027_conference,times}

\usepackage{amsmath,amsfonts,bm}

\def\eqref#1{equation~\ref{#1}}

\def\1{\bm{1}}

\DeclareMathAlphabet{\mathsfit}{\encodingdefault}{\sfdefault}{m}{sl}
\SetMathAlphabet{\mathsfit}{bold}{\encodingdefault}{\sfdefault}{bx}{n}

\usepackage{hyperref}
\hypersetup{hidelinks}
\usepackage{url}
\usepackage{booktabs}
\usepackage{multirow}
\usepackage{tabularx}
\usepackage{array}
\usepackage{makecell}
\usepackage{amssymb}
\usepackage[table]{xcolor}
\usepackage{graphicx}
\usepackage{pifont}
\usepackage{float}

\definecolor{RigidTint}{RGB}{255,248,205}
\definecolor{DeformTint}{RGB}{220,250,222}
\definecolor{ParticleTint}{RGB}{255,234,211}
\definecolor{CheckGreen}{RGB}{46,139,87}
\definecolor{CrossRed}{RGB}{200,65,65}

\newcolumntype{Y}{>{\centering\arraybackslash}X}

\newcommand{\bestcell}[1]{\textbf{#1}}
\newcommand{\cmark}{\textcolor{CheckGreen}{\ding{51}}}
\newcommand{\xmark}{\textcolor{CrossRed}{\ding{55}}}

\title{OneWorld: Learning Consistent Physics\\ Across Actions in World Models}

\author{
Ke He \qquad Yichen Ding \qquad Bin Yang \\
Wuhan University \\
Wuhan, China \\
\texttt{\{koendrunk,dingyichen,yangbin\_cv\}@whu.edu.cn}
}

\iclrfinalcopy

\begin{document}

\maketitle
\lhead{}

\begin{abstract}

Action-conditioned video world models aim to predict scene evolution under different actions, a capability that is essential for reliable planning, decision-making, and interaction in dynamic environments. However, futures generated independently from the same initial scene may each appear plausible while implying incompatible physical properties, such as friction or mass. This inconsistency can lead to contradictory predictions across interventions, making it difficult for the model to maintain a coherent understanding of the underlying world and limiting its reliability for planning and decision-making. To address these issues, we propose \textbf{OneWorld}, a shared-mechanism counterfactual generation framework that jointly models multiple action-conditioned futures under a common latent physical mechanism. A physical mechanism interpreter first infers a distribution over latent mechanisms from each action–outcome branch. These distributions are then aggregated into shared-world evidence, which captures whether the branches admit a common physical explanation while accounting for uncertainty in less informative branches. This evidence constrains flow training and guides sampling, encouraging consistency in the underlying physical mechanism while preserving the distinct outcomes induced by different actions. We further introduce a multi-intervention evaluation protocol in controlled environments, following the interaction settings of ACWM-Phys, to assess whether generated futures can be jointly explained by the same physical parameters, alongside standard measures of single-rollout prediction quality. Experiments in these environments show that OneWorld improves cross-intervention physical consistency while maintaining competitive single-rollout prediction quality.
\textit{\color{magenta}The source code will be released.}
\end{abstract}

\section{Introduction}
\label{sec:introduction}

Action-conditioned video world models aim to predict how a scene evolves
under different actions, a capability that is increasingly important for
planning, decision-making, and interaction in dynamic
environments~\citep{hansen2022tdmpc,yang2024unisim}.
Given an initial visual observation and a candidate action sequence, these
models predict the resulting future video, allowing an agent to compare
possible outcomes before choosing how to act~\citep{huang2026vid2world,xue2026acwmphys}.
Such comparisons implicitly assume that alternative futures correspond to
different interventions on the same underlying world. The actions may change,
and their consequences should differ, but action-independent physical
properties such as object mass, surface friction, or material stiffness
should remain consistent across the predicted futures.

\begin{figure}[t]
    \centering
    \includegraphics[width=0.84\textwidth]{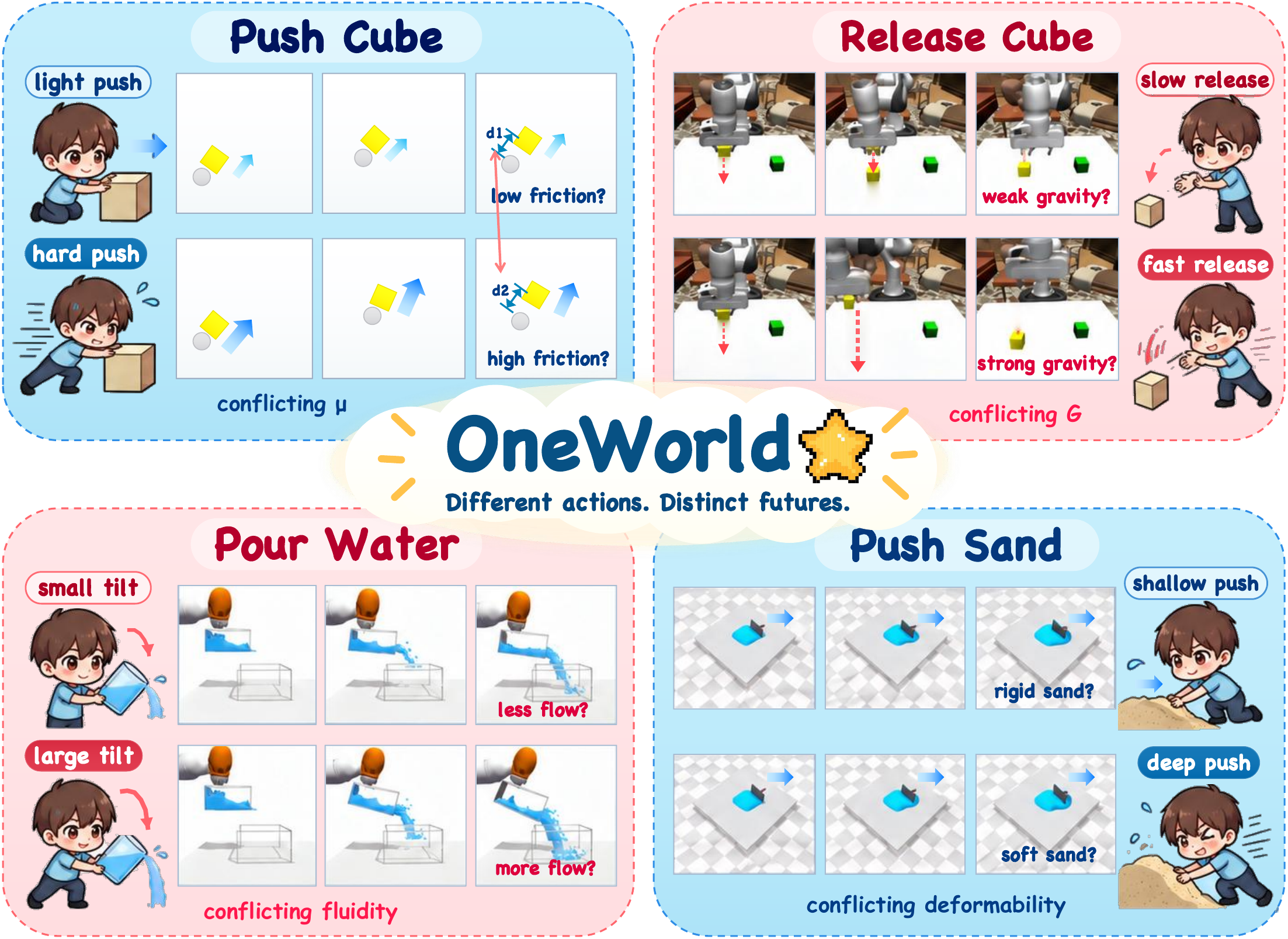}
    \caption{
    \textbf{Motivating probes of action-conditioned prediction.}
    ACWM-DiT predictions under different actions from the same initial
    observation within each scene, illustrating the need to evaluate whether
    alternative futures remain compatible with a shared physical world.
    }
    \label{fig:motivation}
\end{figure}

Current action-conditioned world models do not explicitly enforce consistency
across alternative futures from the same initial world. They typically generate
each future independently under its corresponding action
sequence~\citep{bruce2024genie,guo2026ctrlworld,xue2026acwmphys}. As a result,
several predictions can each appear plausible in isolation while implying
mutually incompatible physical properties. For example, two predicted cube
trajectories under different pushes may each admit a reasonable explanation
individually, yet one may require a high-friction surface while the other is
only compatible with a low-friction one. Figure~\ref{fig:motivation}
illustrates this failure mode by probing ACWM-DiT under multiple actions from
fixed initial observations. We refer to this phenomenon as
\emph{shared-world inconsistency}. Our goal is to achieve
\emph{shared-world consistency}, where action-dependent futures generated from
the same initial world admit a common physical explanation while preserving
their distinct trajectories. To quantify this property, we compare separate
branch-wise physical fits with a joint fit that requires all branches to share
one physical configuration. The additional fitting error introduced by this
shared constraint defines the \emph{shared-world gap}. The nonzero gaps
observed for ACWM-DiT in Table~\ref{tab:main_results} show that individually
plausible rollouts can remain inconsistent when interpreted as outcomes of one
shared physical world.

Modeling shared-world consistency introduces two key challenges.
First, the physical properties implied by a prediction cannot be inferred
from the outcome alone; they must be interpreted jointly with the action that
produced it. The displacement of a pushed object, for instance, depends on
both the applied action and the physical properties of the object and
surface. Second, different interventions provide unequal amounts of
information about those properties. A branch in which an object is barely
disturbed may remain compatible with many physical configurations, whereas
a strong interaction can sharply constrain the plausible mechanisms.
Consequently, physically consistent branches need not induce identical
beliefs about the world: they may support the same mechanism with different
levels of uncertainty.

To address these challenges, we propose \textbf{OneWorld}, a
shared-mechanism counterfactual generation framework that jointly models
multiple action-conditioned futures through a common latent physical
mechanism. A \emph{physical mechanism interpreter} first maps each
action--outcome branch to a distribution over latent mechanisms.
We then introduce a \emph{shared-world evidence score} that aggregates these
branch distributions relative to an initial-scene prior. The score measures
whether the branches retain common support for a shared mechanism while
allowing less informative interventions to remain uncertain. We first train
the interpreter using same-world and mismatched-world groups, and then freeze
it to provide a physical compatibility signal for both flow training and
sampling-time guidance. In this way, OneWorld couples the physical
interpretations of different branches without forcing their predicted videos
or mechanism posteriors to become identical.

We evaluate OneWorld on controlled Push Cube, Push Rope, and Pour Water
environments following the corresponding interaction settings of
ACWM-Phys~\citep{xue2026acwmphys}. These environments span rigid-body,
deformable-object, and particle-based dynamics and allow multiple
interventions to be executed from the same underlying physical
configuration. We evaluate both conventional single-rollout prediction
quality and whether the resulting futures can be jointly explained by one
physical configuration. Relative to ACWM-DiT, OneWorld reduces the mean
shared-world gap by \textbf{84.3\%} while improving mean M-MSE by
approximately $7.7\times10^{-4}$.

We make the following contributions.
\begin{itemize}
    \item We formulate \emph{shared-world consistency}: action-dependent
    futures generated from one initial world should admit a common physical
    explanation.

    \item We introduce a physical mechanism interpreter and a prior-corrected
    shared-world evidence score that assess compatibility across interventions
    while accounting for unequal uncertainty.

    \item We use the shared-world evidence signal to couple
    action-conditioned predictions through shared-world flow training and
    sampling-time flow guidance.

    \item We introduce an independent simulator-based evaluation protocol
    that separates individual physical fit from cross-intervention
    shared-world consistency, and demonstrate substantial consistency gains
    across three controlled physical environments.
\end{itemize}

\section{Related Work}
\label{sec:related_work}

\paragraph{Action-conditioned video world models.}
Diffusion and flow-matching models provide the generative foundation for
high-capacity visual prediction~\citep{ho2020ddpm,rombach2022ldm,ho2022videodiffusion,lipman2023flowmatching}.
World models extend such predictors to planning and interaction by conditioning
future observations on actions~\citep{hansen2024tdmpc2,bruce2024genie,huang2026vid2world,guo2026ctrlworld}.
Recent systems increasingly model long-horizon, visually rich dynamics, but
their training objectives still primarily supervise the quality of each
action-conditioned rollout individually.

\paragraph{Physical reasoning and evaluation.}
Physical priors and post-training objectives have been introduced to improve
physical behavior in generated videos~\citep{wang2025wisa,yuan2026newtongen,wang2026prophy},
while benchmarks evaluate whether generated motion follows plausible physical
dynamics~\citep{bansal2025videophy,motamed2026physicsiq,xue2026acwmphys}.
ACWM-Phys provides the interaction settings and action-conditioned backbone
used in our controlled experiments. Existing evaluation is largely
single-rollout, whereas our protocol asks whether several futures from the same
initial world can be explained by one shared physical configuration.

\paragraph{Counterfactual and multi-branch consistency.}
CoPhy reasons about latent physical factors for counterfactual
prediction~\citep{baradel2020cophy}, and Multiverse Mechanica studies causal
consistency across parallel worlds with matched initial
conditions~\citep{ness2026multiverse}. CoCo improves action responsiveness
through counterfactual constraints~\citep{shi2026coco}, while Twin Rollouts
couples factual and counterfactual branches through shared prefixes and future
exogenous noise~\citep{ma2026twinrollouts}. OneWorld instead couples multiple
action-conditioned futures through the compatibility of their inferred
physical explanations, without requiring identical branch posteriors or shared
sampling noise.

\section{Method}
\label{sec:method}

\begin{figure}[H]
    \centering
    \includegraphics[width=\linewidth]{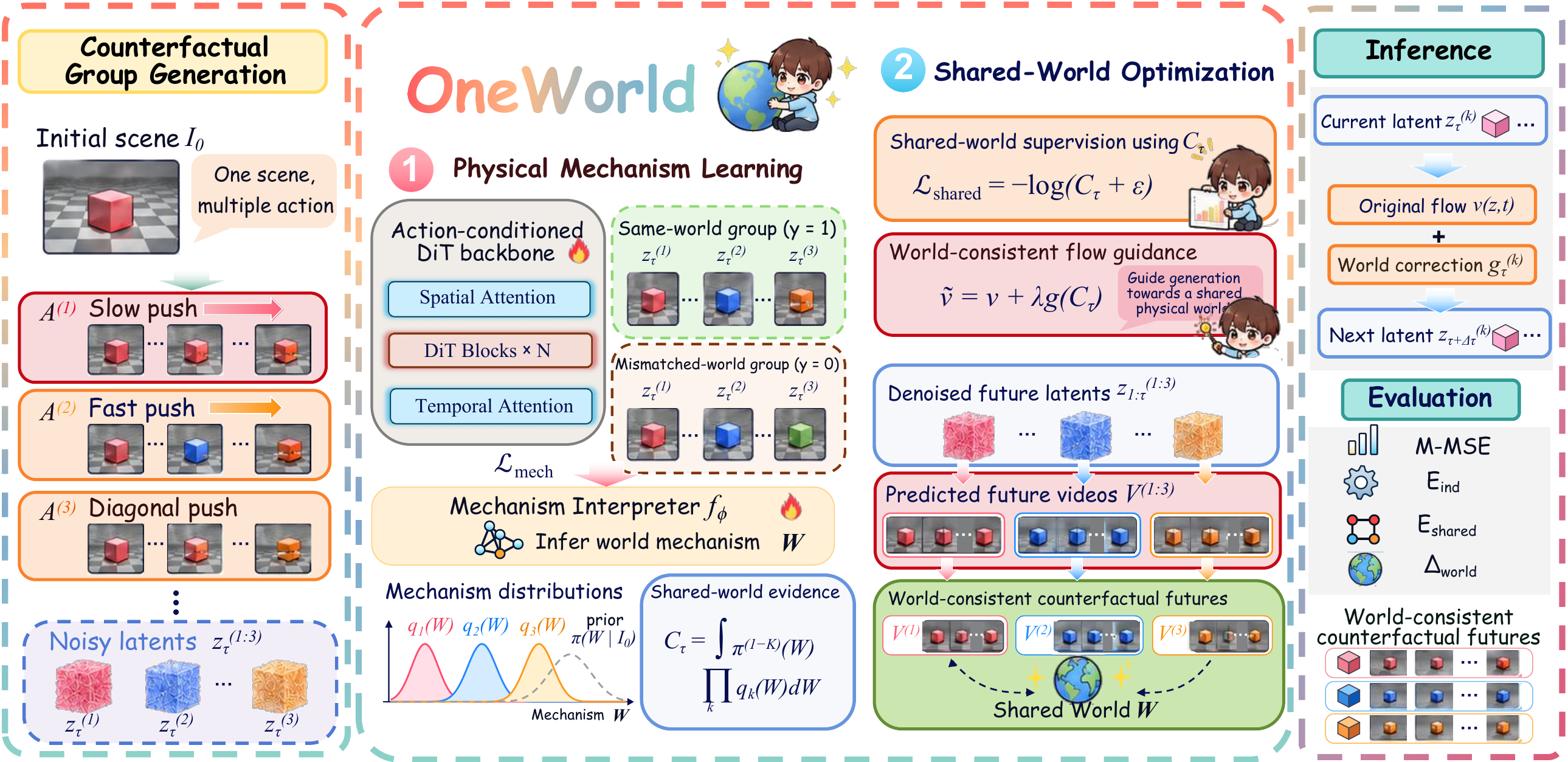}
    \caption{
    Overview of \textbf{OneWorld}. Given the same initial scene and multiple action
    sequences, OneWorld predicts a future for each intervention branch.
    A physical mechanism interpreter infers a mechanism distribution from each
    action--outcome pair. The shared-world evidence score assesses their
    compatibility and provides the signal for shared-world flow training
    and shared-world flow guidance.
    }
    \label{fig:oneworld_overview}
\end{figure}

OneWorld predicts multiple action-conditioned futures from one initial scene
and couples them through a learned measure of physical compatibility
(Figure~\ref{fig:oneworld_overview}).
We first define the shared-world task and the underlying flow model
(Section~\ref{sec:problem}), then construct the mechanism interpreter and
shared-world evidence score (Section~\ref{sec:mechanism}).
We learn the interpreter from same-world and mismatched-world groups
(Section~\ref{sec:mechanism_learning}), freeze it, and use the score for
flow training (Section~\ref{sec:world_flow}) and sampling-time guidance
(Section~\ref{sec:flow_guidance}).

\subsection{Task Formulation and Flow Preliminaries}
\label{sec:problem}

Let $I_0$ denote the initial observation of a scene.
We consider $K$ action interventions from the same initial world, each
defining an intervention branch.
For branch $k$, $A^{(k)}=\{a_t^{(k)}\}_{t=0}^{T-1}$ is the action sequence
and $V^{(k)}=\{I_t^{(k)}\}_{t=1}^{T}$ is the corresponding future video.
We refer to these $K$ branches collectively as a \emph{counterfactual group}.
We use $W$ to denote a latent mechanism variable representing
action-independent properties of the world, such as those associated
with friction, mass, or stiffness.
Its coordinates need not correspond to named physical quantities.
We distinguish $W$ from the explicit simulator parameters $w$ used for
evaluation in Section~\ref{sec:exp_setup}.
Different interventions are expected to produce different outcomes, but all
branches originating from the same initial world should be compatible with the
same realization of $W$.
This requirement motivates the following shared-mechanism factorization
\begin{equation}
p\!\left(
V^{(1:K)}
\mid
I_0,A^{(1:K)}
\right)
=
\int
p(W\mid I_0)
\prod_{k=1}^{K}
p\!\left(
V^{(k)}
\mid
I_0,A^{(k)},W
\right)
\,dW ,
\label{eq:joint_world}
\end{equation}
where $A^{(1:K)}$ and $V^{(1:K)}$ denote all intervention sequences and their
future videos.
Equation~\ref{eq:joint_world} assumes conditional independence of the futures
given $I_0$, their actions, and a shared $W$.
For the pushing example, the trajectories under different actions should be compatible with a common physical configuration, including properties such as friction and mass.

\paragraph{Action-conditioned flow model.}
We build OneWorld on an action-conditioned latent flow model following
ACWM-DiT.
Rather than directly parameterizing the conditional factors in
Eq.~\ref{eq:joint_world}, we use an auxiliary mechanism interpreter to
encourage shared-world consistency during flow training and sampling.
Let $z_{\tau}^{(k)}$ denote the future latent of branch $k$ at flow time
$\tau\in[0,1]$, where $\tau=0$ corresponds to noise and $\tau=1$ to the clean
future.
For a linear flow path,
$z_{\tau}^{(k)}=(1-\tau)\epsilon^{(k)}+\tau z_1^{(k)}$, where
$\epsilon^{(k)}$ is Gaussian noise and $z_1^{(k)}$ is the clean future latent.
The pretrained action-conditioned model predicts a velocity
$v_{\theta}(z_{\tau}^{(k)},\tau,I_0,A^{(k)})$ for each branch.
OneWorld keeps this branch-specific action prediction while introducing
cross-branch constraints on the physical mechanism.

At flow time $\tau$, the current vector field provides an estimate of the clean
future latent,
\begin{equation}
\hat{z}_{1,\tau}^{(k)}
=
z_{\tau}^{(k)}
+
(1-\tau)
v_{\theta}
\left(
z_{\tau}^{(k)},
\tau,
I_0,
A^{(k)}
\right).
\label{eq:clean_estimate}
\end{equation}
Under the linear flow path, Eq.~\ref{eq:clean_estimate} recovers the clean
future when the predicted velocity is exact.
We use this estimate rather than the noisy latent itself because physical
properties are expressed through the predicted outcome of the intervention.

\subsection{Mechanism Interpretation and Shared-World Evidence}
\label{sec:mechanism}

\paragraph{Interpreting an action--outcome pair.}
We infer the physical properties implied by a prediction jointly with the
action that produced it. Using the clean-future estimate in
Eq.~\ref{eq:clean_estimate}, a physical mechanism interpreter $q_{\phi}$ maps
each branch to a distribution over the latent mechanism $W$,
\begin{equation}
q_{\phi}^{(k)}(W)
=
q_{\phi}
\left(
W
\mid
I_0,
A^{(k)},
\hat{z}_{1,\tau}^{(k)},
\tau
\right).
\label{eq:mechanism_posterior}
\end{equation}
Conditioning on the action disambiguates similar motions produced by different
interventions, while $\tau$ allows uncertainty to reflect the reliability of
the current future estimate.

\paragraph{Assessing compatibility across branches.}
Let $\pi_{\phi}(W\mid I_0)$ denote the mechanism prior before observing any
intervention outcome. We define the shared-world evidence of a counterfactual
group as
\begin{equation}
\mathcal{C}_{\tau}
=
\int
\pi_{\phi}(W\mid I_0)^{1-K}
\prod_{k=1}^{K}
q_{\phi}^{(k)}(W)
\,dW .
\label{eq:shared_evidence}
\end{equation}
A high $\mathcal{C}_{\tau}$ indicates common posterior support relative to the
initial-scene prior. The factor $\pi_{\phi}^{1-K}$ removes repeated prior
contributions from the branch posteriors. Under exact Bayesian posteriors and
conditional independence, Eq.~\ref{eq:shared_evidence} becomes an evidence
ratio between a shared-mechanism model and independently drawn mechanisms
(Appendix~\ref{app:evidence}); here it is used as a learned compatibility score.

\paragraph{Unequal information across interventions.}
If an uninformative branch satisfies
$q_{\phi}^{(j)}(W)=\pi_{\phi}(W\mid I_0)$, its contribution cancels one prior
factor and leaves the score for the remaining $K-1$ branches unchanged
(Appendix~\ref{app:prior_neutrality}). We implement this behavior by
parameterizing each branch posterior as an additive evidence update to the
initial-scene prior in Gaussian natural-parameter space. Weak interventions
contribute little additional precision, whereas informative interventions
produce sharper mechanism estimates. The parameterization and numerical
estimation are detailed in Appendix~\ref{app:implementation}.

\subsection{Learning the Mechanism Interpreter}
\label{sec:mechanism_learning}

We train the physical mechanism interpreter using relational supervision
rather than explicit physical-parameter labels. A same-world group contains
different action branches generated from the same initial state and hidden
physical configuration, whereas a mismatched-world group keeps the visible
initial scene matched but combines branches from different physical
configurations. This encourages the interpreter to distinguish variation
caused by actions from variation caused by the underlying world.

Let $\mathcal B$ denote a counterfactual group, with $y=1$ for same-world
groups and $y=0$ for mismatched-world groups. Using
$s(\mathcal B)=\log\mathcal C_{\tau}$ as the compatibility score, we optimize
\begin{equation}
\mathcal{L}_{\mathrm{mech}}
=
-
\mathbb{E}_{\mathcal{B}}
\left[
y\log\sigma\!\left(s(\mathcal{B})\right)
+
(1-y)
\log
\left(
1-\sigma\!\left(s(\mathcal{B})\right)
\right)
\right],
\label{eq:mechanism_loss}
\end{equation}
where $\sigma$ is the sigmoid. The supervision acts only at the group level,
so the latent coordinates are not required to correspond to named physical
quantities.

The interpreter and initial-scene prior are trained before the video model.
During this stage, ACWM-DiT is frozen and $\tau$ is sampled across the flow
trajectory, exposing the interpreter to clean-future estimates with different
levels of uncertainty. After training, the interpreter and prior are fixed and
used as the compatibility criterion for shared-world flow training and
guidance. Further details are given in Appendix~\ref{app:training_schedule}.

\subsection{Shared-World Flow Training}
\label{sec:world_flow}

After mechanism learning, we freeze both the interpreter and the initial-scene
prior so that the compatibility criterion remains fixed while the video model
is updated. For each same-world group, the branch-wise clean-future estimates
are interpreted jointly and optimized to yield high shared-world evidence,
\begin{equation}
\mathcal{L}_{\mathrm{shared}}
=
-
\mathbb{E}
\left[
\log
\left(
\mathcal{C}_{\tau}
+
\epsilon
\right)
\right],
\label{eq:shared_loss}
\end{equation}
where $\epsilon$ ensures numerical stability. This loss couples the branches
through their physical interpretations rather than by directly matching their
videos, latents, or posterior parameters.

The complete video-model objective is
\begin{equation}
\mathcal{L}
=
\mathcal{L}_{\mathrm{flow}}
+
\lambda_{\mathrm{shared}}
\mathcal{L}_{\mathrm{shared}},
\label{eq:total_loss}
\end{equation}
where branch-wise flow matching $\mathcal L_{\mathrm{flow}}$ preserves
supervision for the action-specific future, while
$\mathcal L_{\mathrm{shared}}$ discourages sets of predictions that cannot be
explained by one common mechanism. Freezing the interpreter prevents the
compatibility model from adapting to the generator during this stage, but does
not block optimization of the video model. Gradients from the shared-world
loss pass through the interpreter inputs and the clean-future estimate in
Eq.~\ref{eq:clean_estimate} to the action-conditioned vector field.

\subsection{Shared-World Flow Guidance}
\label{sec:flow_guidance}

Shared-world training improves the learned vector field, but individual
sampling trajectories can still drift toward futures whose inferred mechanisms
are incompatible. We therefore reuse the frozen evidence model at inference
time. For branch $k$, we compute
\begin{equation}
g_{\tau}^{(k)}
=
\nabla_{z_{\tau}^{(k)}}
\log
\left(
\mathcal{C}_{\tau}
+
\epsilon
\right),
\label{eq:world_gradient}
\end{equation}
which gives the local latent-space direction that most increases the
compatibility of the current counterfactual group. Because
$\mathcal C_\tau$ depends jointly on all branch posteriors, the correction for
one branch is informed by the physical explanation supported by the others.

We modify the original action-conditioned velocity as
\begin{equation}
\widetilde{v}_{\theta}^{(k)}
=
v_{\theta}
\left(
z_{\tau}^{(k)},
\tau,
I_0,
A^{(k)}
\right)
+
\lambda_{\tau}
g_{\tau}^{(k)},
\label{eq:guided_velocity}
\end{equation}
where $\lambda_{\tau}$ controls the strength of the shared-world correction.
The original velocity remains responsible for action-conditioned prediction,
while the guidance term provides a comparatively small group-level adjustment.
We apply weak guidance early in the trajectory, when clean-future estimates
are uncertain, and increase its influence as generation becomes more
informative. The schedule and gradient approximation are given in
Appendix~\ref{app:gradient_details}. At inference time, branches share only the
initial observation and compatibility signal; their actions and Gaussian noise
remain independent, so the method does not force identical futures or
identical mechanism posteriors.

\section{Experiments}
\label{sec:experiments}

We evaluate shared-world consistency together with single-rollout prediction
quality, compare against alternative coupling strategies, and study the
contributions and sensitivity of the proposed components.

\subsection{Experimental Setup}
\label{sec:exp_setup}

\paragraph{Benchmark.}
Following ACWM-Phys interaction settings~\citep{xue2026acwmphys}, we evaluate
on Push Cube, Push Rope, and Pour Water, covering rigid-body, deformable, and
fluid dynamics. Same-world groups execute different actions from the same
initial state and hidden physical configuration, while mismatched-world groups
keep the visible initial scene matched but vary the hidden configuration.
Full construction details are in Appendix~\ref{app:environment_construction}.

\paragraph{Baselines.}
We compare with ACWM-DiT~\citep{xue2026acwmphys}, Vid2World~\citep{huang2026vid2world},
CoCo~\citep{shi2026coco}, Twin Rollouts~\citep{ma2026twinrollouts}, Shared
Noise, and Posterior Alignment. These respectively represent independent
prediction, action-conditioned video modeling, counterfactual action
constraints, noise-coupled counterfactual branches, shared randomness, and
direct alignment of mechanism posteriors. External objectives are instantiated
on the same action-conditioned DiT backbone, with matched training groups,
actions, optimization budgets, and evaluation protocols.

\paragraph{Evaluation metrics.}
We evaluate both single-rollout prediction quality and shared-world
consistency. For single-rollout prediction, we use Masked-MSE (M-MSE) from
ACWM-Phys~\citep{xue2026acwmphys}, which emphasizes temporally changing,
action-affected regions. For shared-world evaluation, a frozen state extractor
maps each generated video to an observable trajectory $\hat S^{(k)}$.
Replaying the corresponding action from the same initial state under candidate
simulator parameters $w$ gives $S^{(k)}(w)$. Using a normalized
environment-specific state distance $d$, we define
\begin{align}
\ell_k(w) &= \frac{1}{T}\sum_{t=1}^{T}
d\!\left(\hat S_t^{(k)},S_t^{(k)}(w)\right),
\label{eq:trajectory_fit}\\
E_{\mathrm{ind}} &= \frac{1}{K}\sum_{k=1}^{K}\min_{w_k}\ell_k(w_k),
\label{eq:individual_fit}\\
E_{\mathrm{shared}} &= \min_w\frac{1}{K}\sum_{k=1}^{K}\ell_k(w),
\qquad
\Delta_{\mathrm{world}}
=
E_{\mathrm{shared}}-E_{\mathrm{ind}}.
\label{eq:shared_fit}
\end{align}
Here $E_{\mathrm{ind}}$ measures branch-wise physical fit,
$E_{\mathrm{shared}}$ requires one physical configuration to explain all
branches jointly, and $\Delta_{\mathrm{world}}$ measures the additional cost
of enforcing a common world. We therefore report all three metrics jointly to
distinguish shared-world consistency from overall physical fit quality. The
simulator-based evaluator is independent of the learned compatibility score;
further details are provided in Appendix~\ref{app:shared-eval}.

\paragraph{Implementation details.}
We use a frozen Wan2.1 VAE~\citep{wan2025wan}, initialize from ACWM-DiT, and
use $K=3$ branches by default. The interpreter and prior are trained first and
then frozen. Full architecture, optimization, and sampling settings are in
Appendix~\ref{app:implementation}.

\subsection{Main Results}
\label{sec:main_results}

Table~\ref{tab:main_results} compares OneWorld with independent prediction
and alternative cross-branch coupling strategies. We jointly report
single-rollout prediction quality, individual physical fit, and shared-world
consistency.

\begin{table}[t]
\centering
\caption{
Multi-intervention comparison: individual versus shared physical fits
($E_{\mathrm{ind}}$, $E_{\mathrm{shared}}$) and their gap
$\Delta_{\mathrm{world}}$. Values are mean $\pm$ standard deviation over
three independent runs; lower is better for all metrics.
}
\label{tab:main_results}

\scriptsize
\setlength{\tabcolsep}{2.0pt}
\renewcommand{\arraystretch}{1.06}

\resizebox{\linewidth}{!}{%
\begin{tabular}{
@{}
l
l
l
c
c
c
c
@{}
}

\toprule
\multirow{2}{*}{\textbf{Category}}
& \multirow{2}{*}{\textbf{Environment}}
& \multirow{2}{*}{\textbf{Method}}
& \multicolumn{1}{c}{\textbf{Prediction}}
& \multicolumn{3}{c}{\textbf{Physical fit and consistency}}
\\
\cmidrule(lr){4-4}
\cmidrule(lr){5-7}

& &
& \textbf{M-MSE $\downarrow$}
& \textbf{$E_{\mathrm{ind}}\downarrow$}
& \textbf{$E_{\mathrm{shared}}\downarrow$}
& \textbf{$\Delta_{\mathrm{world}}\downarrow$}
\\

\midrule

\multirow{7}{*}{\makecell[l]{\textbf{Rigid}\\\textbf{body}}}
& \multirow{7}{*}{\colorbox{RigidTint}{\strut\textbf{Push Cube}}}
& ACWM-DiT
& $0.03477\pm0.00034$
& $0.07535\pm0.00112$
& $0.09788\pm0.00173$
& $0.02253\pm0.00109$ \\

& & Vid2World
& $0.03428\pm0.00029$
& $0.07488\pm0.00096$
& $0.09346\pm0.00161$
& $0.01858\pm0.00094$ \\

& & CoCo
& \underline{$0.03416\pm0.00031$}
& \underline{$0.07452\pm0.00088$}
& $0.08891\pm0.00139$
& $0.01439\pm0.00082$ \\

& & Twin Rollouts
& $0.03435\pm0.00027$
& $0.07497\pm0.00104$
& $0.09078\pm0.00148$
& $0.01581\pm0.00087$ \\

& & Shared Noise
& $0.03464\pm0.00036$
& $0.08073\pm0.00121$
& $0.11122\pm0.00196$
& $0.03049\pm0.00128$ \\

& & Posterior Alignment
& $0.03486\pm0.00033$
& $0.07547\pm0.00091$
& \underline{$0.08836\pm0.00132$}
& \underline{$0.01289\pm0.00076$} \\

& & \textbf{OneWorld}
& $\bm{0.03396\pm0.00024}$
& $\bm{0.07384\pm0.00079}$
& $\bm{0.07741\pm0.00108}$
& $\bm{0.00357\pm0.00038}$ \\

\midrule

\multirow{7}{*}{\textbf{Deformable}}
& \multirow{7}{*}{\colorbox{DeformTint}{\strut\textbf{Push Rope}}}
& ACWM-DiT
& $0.00817\pm0.00011$
& $0.04128\pm0.00062$
& $0.05517\pm0.00091$
& $0.01389\pm0.00057$ \\

& & Vid2World
& $0.00798\pm0.00009$
& $0.04096\pm0.00055$
& $0.05302\pm0.00083$
& $0.01206\pm0.00049$ \\

& & CoCo
& \underline{$0.00791\pm0.00008$}
& \underline{$0.04072\pm0.00051$}
& $0.05088\pm0.00076$
& $0.01016\pm0.00044$ \\

& & Twin Rollouts
& $0.00802\pm0.00010$
& $0.04105\pm0.00058$
& $0.05214\pm0.00079$
& $0.01109\pm0.00047$ \\

& & Shared Noise
& $0.00809\pm0.00012$
& $0.04311\pm0.00067$
& $0.05972\pm0.00098$
& $0.01661\pm0.00063$ \\

& & Posterior Alignment
& $0.00823\pm0.00009$
& $0.04147\pm0.00053$
& \underline{$0.04961\pm0.00071$}
& \underline{$0.00814\pm0.00041$} \\

& & \textbf{OneWorld}
& $\bm{0.00771\pm0.00007}$
& $\bm{0.04012\pm0.00046}$
& $\bm{0.04221\pm0.00064}$
& $\bm{0.00209\pm0.00026}$ \\

\midrule

\multirow{7}{*}{\textbf{Particle}}
& \multirow{7}{*}{\colorbox{ParticleTint}{\strut\textbf{Pour Water}}}
& ACWM-DiT
& $0.03624\pm0.00039$
& $0.09037\pm0.00134$
& $0.12541\pm0.00217$
& $0.03504\pm0.00146$ \\

& & Vid2World
& $0.03584\pm0.00035$
& $0.08993\pm0.00121$
& $0.12072\pm0.00198$
& $0.03079\pm0.00132$ \\

& & CoCo
& \underline{$0.03567\pm0.00032$}
& \underline{$0.08951\pm0.00116$}
& $0.11564\pm0.00184$
& $0.02613\pm0.00118$ \\

& & Twin Rollouts
& $0.03596\pm0.00036$
& $0.09005\pm0.00127$
& $0.11822\pm0.00191$
& $0.02817\pm0.00124$ \\

& & Shared Noise
& $0.03605\pm0.00041$
& $0.09314\pm0.00142$
& $0.13291\pm0.00231$
& $0.03977\pm0.00157$ \\

& & Posterior Alignment
& $0.03635\pm0.00034$
& $0.09058\pm0.00119$
& \underline{$0.11231\pm0.00172$}
& \underline{$0.02173\pm0.00106$} \\

& & \textbf{OneWorld}
& $\bm{0.03521\pm0.00028}$
& $\bm{0.08846\pm0.00103}$
& $\bm{0.09402\pm0.00153}$
& $\bm{0.00556\pm0.00061}$ \\

\bottomrule
\end{tabular}%
}
\end{table}

\raggedbottom

\paragraph{Overall performance.}
OneWorld consistently improves shared-world consistency across all three
dynamics regimes. Relative to ACWM-DiT, the mean
$\Delta_{\mathrm{world}}$ decreases from $0.02382$ to $0.00374$, corresponding
to an \textbf{84.3\%} reduction. At the same time,
$E_{\mathrm{ind}}$ remains low and $E_{\mathrm{shared}}$ decreases
substantially, indicating that the smaller gap does not arise from uniformly
poor physical fits. OneWorld also achieves the lowest M-MSE in all three
environments, improving the mean by approximately $7.7\times10^{-4}$ over
ACWM-DiT. Additional analyses of ground-truth physical-configuration recovery
and interpolation to unseen physical values are provided in
Appendix~\ref{app:physics_recovery} and
Appendix~\ref{app:physics_interpolation}, respectively.

\paragraph{Comparison with alternative coupling strategies.}
The baselines show that improving action-conditioned prediction alone does not
eliminate shared-world inconsistency. Vid2World, CoCo, and Twin Rollouts
reduce the gap relative to independent ACWM-DiT generation, but remain
substantially above OneWorld. Shared Noise slightly improves M-MSE while
increasing the shared-world gap, indicating that common randomness is not
sufficient to enforce a common physical world. Posterior Alignment provides
the strongest alternative on the consistency metrics by using the same
mechanism interpreter, yet direct posterior similarity remains markedly
weaker than the prior-corrected shared-support criterion used by OneWorld.

\subsection{Qualitative Analysis on RoboDesk}
\label{sec:qualitative}

\begin{figure}[H]
    \centering
    \includegraphics[width=\linewidth]{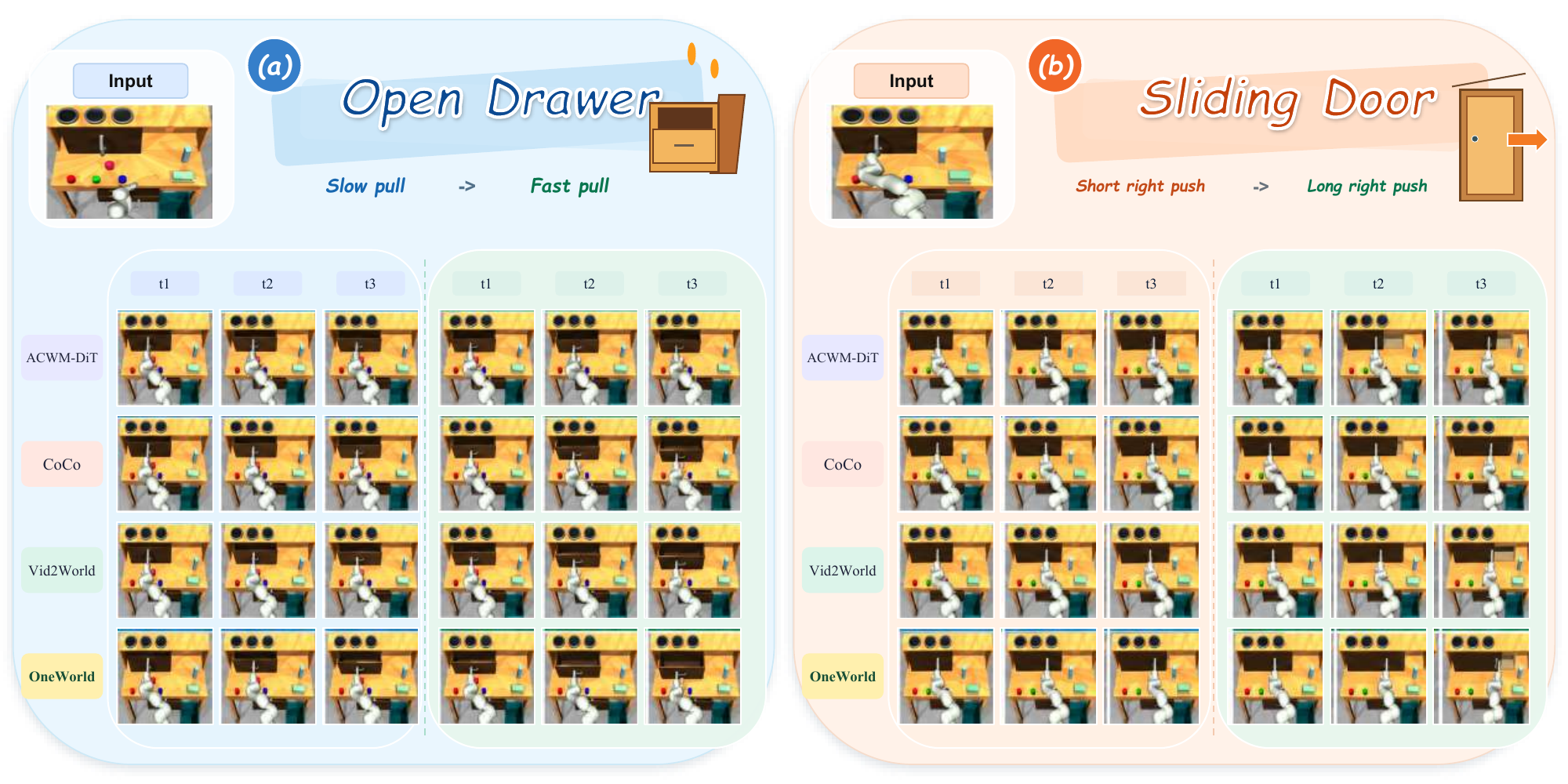}
    \caption{
    \textbf{Different actions from the same initial world on RoboDesk.}
    Branches share the initial state and hidden physics but differ in action.
    Consistency concerns a common physical explanation after accounting for
    contact and release conditions, rather than identical displacements.
    }
    \label{fig:qualitative}
\end{figure}

Figure~\ref{fig:qualitative} compares action responses from the same initial
world in two RoboDesk cases. OneWorld preserves distinct outcomes for slow
versus fast drawer pulls and short versus long door pushes while maintaining
coherent post-contact dynamics across interventions.
Appendix~\ref{app:robodesk_qualitative_protocol} provides the complete setup
and contact/release-conditioned interpretation.

\flushbottom 
\subsection{Ablation Studies}
\label{sec:ablations}

Table~\ref{tab:ablation} isolates three components on \textbf{Push Cube}
with the same backbone and evaluation protocol.

\begin{table}[H]
\centering
\caption{
Ablation of OneWorld.
``Mismatch Neg.'' denotes mismatched-world supervision for the mechanism
interpreter.
``Shared Train'' denotes shared-world flow training with
$\mathcal{L}_{\mathrm{shared}}$, and ``Flow Guide'' denotes shared-world
flow guidance during sampling.
}
\label{tab:ablation}
\small
\setlength{\tabcolsep}{3.0pt}
\renewcommand{\arraystretch}{1.10}

\begin{tabularx}{\linewidth}{
@{}
>{\raggedright\arraybackslash}X
>{\centering\arraybackslash}p{0.12\linewidth}
>{\centering\arraybackslash}p{0.12\linewidth}
>{\centering\arraybackslash}p{0.12\linewidth}
>{\centering\arraybackslash}p{0.125\linewidth}
>{\centering\arraybackslash}p{0.125\linewidth}
>{\centering\arraybackslash}p{0.125\linewidth}
@{}
}

\toprule
\multirow{2}{*}{\textbf{Method}}
& \multicolumn{3}{c}{\textbf{Components}}
& \multicolumn{3}{c}{\textbf{Metrics}}
\\
\cmidrule(lr){2-4}\cmidrule(lr){5-7}
& \makecell{\textbf{Mismatch}\\\textbf{Neg.}}
& \makecell{\textbf{Shared}\\\textbf{Train}}
& \makecell{\textbf{Flow}\\\textbf{Guide}}
& \textbf{M-MSE $\downarrow$}
& \textbf{$E_{\mathrm{shared}}\downarrow$}
& \textbf{$\Delta_{\mathrm{world}}\downarrow$}
\\
\midrule

ACWM-DiT
& --
& --
& --
& 0.03477
& 0.09788
& 0.02253 \\

\makecell[l]{No Mismatch\\Neg.}
& \xmark
& \cmark
& \cmark
& 0.03442
& 0.08495
& 0.01084 \\

\makecell[l]{No Shared\\Train}
& \cmark
& \xmark
& \cmark
& 0.03431
& 0.08260
& 0.00831 \\

\makecell[l]{No Flow\\Guide}
& \cmark
& \cmark
& \xmark
& 0.03418
& 0.08024
& 0.00625 \\

\textbf{OneWorld}
& \cmark
& \cmark
& \cmark
& \textbf{0.03396}
& \textbf{0.07741}
& \bestcell{0.00357} \\

\bottomrule
\end{tabularx}
\end{table}

Removing mismatched-world supervision worsens the gap, supporting its role
in learning compatibility without establishing physical semantics for
individual latent coordinates. Shared-world training and guidance each
reduce the gap on their own; together they give the lowest gap and
$E_{\mathrm{shared}}$ among tested variants. M-MSE remains close across
variants ($0.03396$--$0.03477$).

\begin{figure}[t]
    \centering
    \includegraphics[width=\linewidth]{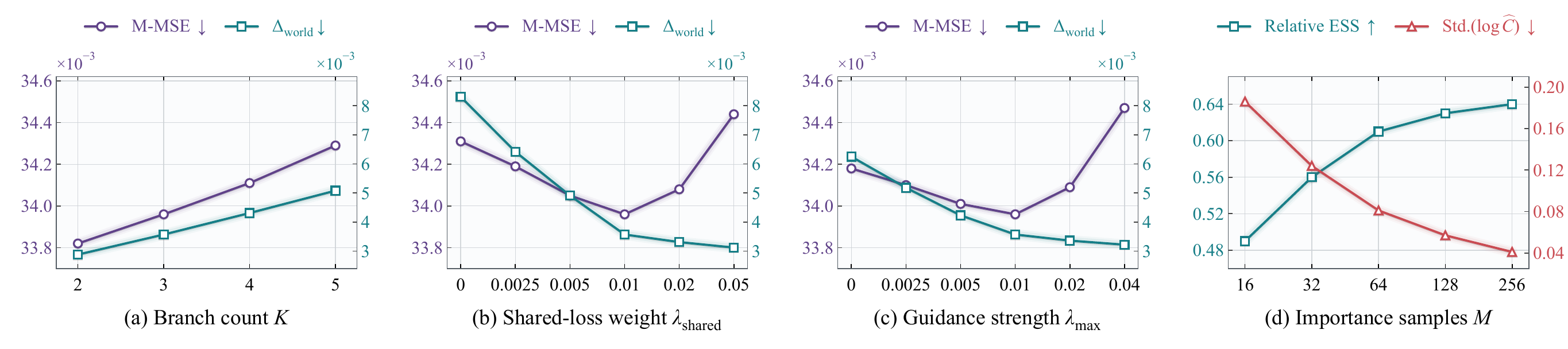}
    \caption{
    Sensitivity of OneWorld to the number of intervention branches $K$,
    shared-loss weight $\lambda_{\mathrm{shared}}$, maximum guidance strength
    $\lambda_{\max}$, and the number of importance samples $M$ on Push Cube.
    }
    \label{fig:sensitivity}
\end{figure}

\paragraph{Sensitivity analysis.}
On Push Cube, increasing $K$ from $2$ to $5$ raises $\Delta_{\mathrm{world}}$
with small M-MSE changes (Figure~\ref{fig:sensitivity}). Both
$\lambda_{\mathrm{shared}}$ and $\lambda_{\max}$ trade consistency against
prediction quality. More importance samples stabilize the estimate, with
limited gains beyond $M=64$; detailed trends are in
Appendix~\ref{app:sensitivity_details}.

\section{Conclusion}
\label{sec:conclusion}

We introduced OneWorld to improve shared-world consistency among action-conditioned futures. A physical mechanism interpreter estimates a latent mechanism distribution for each branch, while the prior-corrected shared-world evidence score couples branches through shared-world flow training and sampling-time guidance. Our simulator-based evaluation separates individual physical fit from the additional error introduced when multiple futures are required to share one physical configuration. Across three controlled environments, OneWorld consistently improves cross-intervention physical consistency while maintaining strong single-rollout prediction quality. Additional recovery and interpolation analyses further show that the resulting shared physical explanations remain aligned with the underlying simulator configurations beyond the discrete settings seen during training. More broadly, the results suggest that evaluating world models only one rollout at a time can overlook inconsistencies that become visible only when alternative interventions are considered jointly. These results highlight shared-world consistency as an important complement to conventional single-rollout evaluation for action-conditioned world models.

\subsection*{AI use statement}

Generative AI tools were used for language editing, manuscript organization,
and \LaTeX{} drafting assistance. All AI-assisted text, equations, citations,
experimental descriptions, and numerical claims were reviewed by the authors.
The authors take responsibility for the correctness and originality of the
final submission.

\subsection*{Ethics statement}

This work studies the consistency of action-conditioned world models in
controlled simulation and on public video-prediction benchmarks. It does not
involve human subjects, personally identifying information, or deployment in a
live robotic system. The simulator-based evaluation is intended to diagnose
model inconsistency rather than to infer physical properties of real
individuals or private environments.

\subsection*{Reproducibility statement}

Section~\ref{sec:method} specifies the shared-world evidence score, mechanism
learning objective, flow-training objective, and sampling-time guidance.
Section~\ref{sec:exp_setup} describes the main evaluation protocol.
Appendix~\ref{app:implementation} provides architecture, optimization, and
sampling details, while Appendix~\ref{app:shared-eval} specifies environment
construction, group formation, state extraction, physical fitting, grid-density
checks, physical-configuration recovery, and interpolation evaluation. We will
release the reconstructed environments, training configurations, evaluator
code, and split definitions with the source code.

\bibliography{iclr2027_conference}
\bibliographystyle{iclr2027_conference}
\appendix


\section{Properties of Shared-World Evidence}
\label{app:evidence}

This appendix provides additional details on the shared-world evidence
defined in Eq.~\ref{eq:shared_evidence}. We first derive its
prior-corrected evidence-ratio interpretation under an idealized Bayesian
setting and then establish the neutrality of a branch whose posterior
contains no information beyond the initial-scene prior. These results
motivate the prior-corrected form of the compatibility score used by
OneWorld.

\subsection{An Idealized Evidence-Ratio Interpretation}
\label{app:evidence_ratio}

Fix the initial observation, intervention sequences, and flow time, and
write $x_k$ for the outcome representation associated with branch $k$.
We suppress this fixed conditioning to simplify notation.

Let $\pi(W)$ be a normalized prior over the latent mechanism and let
\[
L_k(W)=p(x_k\mid W)
\]
denote the likelihood of the outcome of branch $k$ under its corresponding
fixed action. Define the marginal likelihood
\[
Z_k
=
\int \pi(W)L_k(W)\,dW,
\]
which is assumed to be finite and positive.

Under an idealized Bayesian model, the exact branch posterior is
\[
q_k(W)
=
\frac{\pi(W)L_k(W)}{Z_k}.
\]

Substituting the branch posteriors into the shared-world evidence yields
\[
\begin{aligned}
\mathcal{C}
&=
\int
\pi(W)^{1-K}
\prod_{k=1}^{K}
q_k(W)\,dW
\\
&=
\int
\pi(W)^{1-K}
\prod_{k=1}^{K}
\frac{\pi(W)L_k(W)}{Z_k}
\,dW
\\
&=
\frac{
\int
\pi(W)
\prod_{k=1}^{K}L_k(W)\,dW
}{
\prod_{k=1}^{K}Z_k
}.
\end{aligned}
\]

The numerator is the marginal likelihood under a model in which all
branches share a single realization of $W$. The denominator is the product
of the branch-wise marginal likelihoods and corresponds to independently
drawing one mechanism from the same prior for each branch.

The correction factor
\[
\pi(W)^{1-K}
\]
therefore removes the repeated prior factors introduced by multiplying the
$K$ posterior distributions. The shared-mechanism model contains the prior
only once.

Under these assumptions, a larger value of $\mathcal{C}$ represents
stronger relative evidence that the observed action--outcome branches can
be jointly explained by a common mechanism. In OneWorld,
$q_{\phi}$ is learned from relational supervision rather than exact
likelihoods. We therefore use $\mathcal{C}_{\tau}$ as a differentiable
compatibility score and do not interpret its absolute magnitude as a
calibrated Bayes factor.

\subsection{Neutrality of a Prior-Matching Branch}
\label{app:prior_neutrality}

For a set of intervention branches $\mathcal{S}$, define
\[
\mathcal{C}_{\mathcal{S}}
=
\int
\pi_{\phi}(W\mid I_0)^{1-|\mathcal{S}|}
\prod_{k\in\mathcal{S}}
q_{\phi}^{(k)}(W)
\,dW.
\]

Assume the integral is finite and the prior has positive density on the
relevant support. Consider a branch $j\in\mathcal{S}$ whose posterior
contains no additional information beyond the initial observation:
\[
q_{\phi}^{(j)}(W)
=
\pi_{\phi}(W\mid I_0).
\]

Then
\[
\begin{aligned}
\mathcal{C}_{\mathcal{S}}
&=
\int
\pi_{\phi}(W\mid I_0)^{1-|\mathcal{S}|}
q_{\phi}^{(j)}(W)
\prod_{k\in\mathcal{S}\setminus\{j\}}
q_{\phi}^{(k)}(W)
\,dW
\\
&=
\int
\pi_{\phi}(W\mid I_0)^{2-|\mathcal{S}|}
\prod_{k\in\mathcal{S}\setminus\{j\}}
q_{\phi}^{(k)}(W)
\,dW
\\
&=
\mathcal{C}_{\mathcal{S}\setminus\{j\}}.
\end{aligned}
\]

Thus, adding a branch whose posterior equals the prior does not change the
shared-world evidence. For a single branch, the score is $1$ by
normalization, and it also remains $1$ when all branch posteriors equal the
prior.

This property is algebraic. It does not by itself establish that the
learned interpreter produces a prior-matching posterior for every
empirically uninformative intervention. The latter remains a property of
the learned representation rather than of the score definition.


\section{External Action-Conditioned Video Prediction}
\label{app:external_benchmarks}

We additionally evaluate the action-conditioned DiT backbone used by
OneWorld on BAIR and RoboNet using standard video-prediction metrics. These
datasets provide real robot interaction trajectories but not the controlled
multi-intervention physical configurations required by the shared-world
evaluator, so this appendix focuses on single-rollout prediction quality.

\subsection{Evaluation Scope}
\label{app:external_scope}

For both datasets, the external prediction task has the form
\[
p
\left(
V_{\mathrm{future}}
\mid
V_{\mathrm{context}},
A_{\mathrm{future}}
\right),
\]
where the model observes a short visual context together with the future
robot action sequence and predicts the corresponding future video.

For BAIR, we use two observed context frames and predict the following
twelve frames. For RoboNet, we use two observed context frames and predict
the following ten frames. All models compared within each dataset use the
same context length and prediction horizon.

Both datasets are evaluated at a spatial resolution of
$64\times64$.

BAIR and RoboNet are evaluated with conventional single-rollout metrics;
the shared-world metrics are reserved for the controlled environments in the
main experiments.

\subsection{Preprocessing and Temporal Alignment}
\label{app:external_preprocessing}

Frames are resized to $64\times64$ before video encoding. The same
deterministic spatial transformation is applied to all frames from one
trajectory so that preprocessing does not introduce artificial temporal
motion.

RGB values are normalized using the preprocessing convention of the
frozen video autoencoder. No random crop, horizontal flip, color
augmentation, or other test-time augmentation is applied during
evaluation.

Robot actions are temporally aligned with their corresponding target
transitions after frame sampling. Each action vector is standardized using
the mean and standard deviation computed from the training split of the
corresponding dataset.

For BAIR, evaluation uses 2,048 held-out prediction clips. For RoboNet,
evaluation uses 2,048 held-out clips sampled from trajectories that do not
appear in the training split. The same evaluation clips are reused across
all locally evaluated models.

\begin{table}[H]
\centering
\caption{Protocol for external action-conditioned video prediction.}
\label{tab:external_protocol}
\small
\setlength{\tabcolsep}{3.6pt}
\renewcommand{\arraystretch}{1.08}

\begin{tabularx}{\linewidth}{
@{}
>{\raggedright\arraybackslash}p{0.16\linewidth}
>{\centering\arraybackslash}p{0.14\linewidth}
>{\centering\arraybackslash}p{0.14\linewidth}
>{\centering\arraybackslash}p{0.16\linewidth}
>{\centering\arraybackslash}X
@{}
}
\toprule
\textbf{Dataset}
&
\textbf{Resolution}
&
\textbf{Context}
&
\textbf{Prediction}
&
\textbf{Evaluation clips}
\\
\midrule

BAIR
&
$64\times64$
&
2 frames
&
12 frames
&
2,048
\\

RoboNet
&
$64\times64$
&
2 frames
&
10 frames
&
2,048
\\

\bottomrule
\end{tabularx}
\end{table}

\subsection{Model Configuration on BAIR and RoboNet}
\label{app:external_training}

The external experiments use the same action-conditioned latent DiT family
as the main model. A dataset-specific action-conditioned DiT is trained for
each benchmark with the frozen Wan2.1 causal VAE. Because BAIR and RoboNet do
not provide the grouped physical supervision used by OneWorld, training is
restricted to standard branch-wise action-conditioned flow matching; the
mechanism interpreter, shared-world loss, and shared-world guidance are not
used in this appendix.

\begin{table}[H]
\centering
\caption{Training configuration for the external video prediction benchmarks.}
\label{tab:external_training}
\small
\setlength{\tabcolsep}{4pt}
\renewcommand{\arraystretch}{1.07}

\begin{tabularx}{\linewidth}{
@{}
>{\raggedright\arraybackslash}p{0.30\linewidth}
>{\centering\arraybackslash}p{0.29\linewidth}
>{\centering\arraybackslash}X
@{}
}
\toprule
\textbf{Setting}
&
\textbf{BAIR}
&
\textbf{RoboNet}
\\
\midrule

Backbone
&
Action-conditioned DiT
&
Action-conditioned DiT
\\

Video representation
&
Frozen Wan2.1 causal VAE
&
Frozen Wan2.1 causal VAE
\\

Optimizer
&
AdamW
&
AdamW
\\

Learning rate
&
$1\times10^{-5}$
&
$1\times10^{-5}$
\\

Batch size
&
8
&
8
\\

Training updates
&
40,000
&
50,000
\\

Inference steps
&
50
&
50
\\

Mechanism interpreter
&
Not used
&
Not used
\\

Shared-world loss
&
Not used
&
Not used
\\

Shared-world guidance
&
Not used
&
Not used
\\

\bottomrule
\end{tabularx}
\end{table}

\subsection{Baselines and Result Provenance}
\label{app:external_baselines}

The comparison includes recurrent video predictors, hierarchical latent
variable models, masked-token predictors, autoregressive world models,
and diffusion-based approaches.

Specifically, we compare against SVG, GHVAE, FitVid, MaskViT,
iVideoGPT, DPM, DyDiff, and SAMPO when the corresponding metric is
available under the matched action-conditioned protocol.

Baseline values are taken from the corresponding published results or from
the matched-protocol comparison reported by the later method when the
original paper does not contain the complete metric set. We do not combine
results reported at different spatial resolutions in the same table.
The DPM values are the diffusion-baseline results reported together with
DyDiff, which is indicated by the starred venue entry.

Our action-conditioned DiT backbone is evaluated using the protocol
described above. A dash indicates that a metric was not available under a
sufficiently matched action-conditioned setting.

\subsection{Metric Computation}
\label{app:external_metrics}

We report Fr\'echet Video Distance (FVD), peak signal-to-noise ratio
(PSNR), structural similarity (SSIM), and LPIPS.

FVD is computed using I3D video features extracted from 2,048 generated
clips and the corresponding 2,048 reference clips. Generated and reference
videos use identical spatial preprocessing and the same temporal horizon
within each dataset.

PSNR, SSIM, and LPIPS are evaluated over predicted future frames only.
Frame-level scores are first averaged temporally within each predicted
sequence and are then averaged across the evaluation set.

Following the reporting convention of the compared methods, SSIM and
LPIPS values in Table~\ref{tab:external_benchmarks} are multiplied by
$100$.

\subsection{External Prediction Results}

\begin{table}[H]
\centering
\caption{
Comparison with existing action-conditioned video prediction methods on
BAIR and RoboNet at $64\times64$ resolution.
Lower is better for FVD and LPIPS; higher is better for PSNR and SSIM.
SSIM and LPIPS are multiplied by $100$.
$^{*}$DPM denotes the diffusion baseline reported with DyDiff.
}
\label{tab:external_benchmarks}

\scriptsize
\setlength{\tabcolsep}{4pt}
\renewcommand{\arraystretch}{1.07}

\begin{tabularx}{\linewidth}{
>{\raggedright\arraybackslash}p{0.12\linewidth}
>{\raggedright\arraybackslash}p{0.18\linewidth}
>{\centering\arraybackslash}p{0.14\linewidth}
*{4}{>{\centering\arraybackslash}X}
}

\toprule
\textbf{Dataset}
&
\textbf{Method}
&
\textbf{Venue}
&
\textbf{FVD $\downarrow$}
&
\textbf{PSNR $\uparrow$}
&
\textbf{SSIM $\uparrow$}
&
\textbf{LPIPS $\downarrow$}
\\
\midrule

\multirow{6}{*}{\textbf{BAIR}}

& MaskViT
& ICLR'23
& 70.5
& --
& --
& -- \\

& iVideoGPT
& NeurIPS'24
& 60.8
& 24.5
& 90.2
& 5.0 \\

& DPM
& ICLR'25$^{*}$
& 48.5
& 25.9
& 92.0
& 4.5 \\

& DyDiff
& ICLR'25
& \underline{45.0}
& 26.2
& 92.4
& 4.2 \\

& SAMPO
& NeurIPS'25
& 55.5
& \underline{26.7}
& \underline{94.7}
& \underline{3.7} \\

\rowcolor{blue!8}
& \textbf{Our DiT backbone}
& \textbf{Ours}
& \textbf{41.2}
& \textbf{27.9}
& \textbf{95.8}
& \textbf{2.8} \\[-1pt]

\specialrule{0.5pt}{1pt}{1pt}

\multirow{9}{*}{\textbf{RoboNet}}

& MaskViT
& ICLR'23
& 133.5
& 23.2
& 80.5
& 4.2 \\

& SVG
& NeurIPS'19
& 123.2
& 23.9
& 87.8
& 6.0 \\

& GHVAE
& CVPR'21
& 95.2
& 24.7
& 89.1
& 3.6 \\

& FitVid
& arXiv'21
& 62.5
& 28.2
& 89.3
& \underline{2.4} \\

& iVideoGPT
& NeurIPS'24
& 63.2
& 27.8
& 90.6
& 4.9 \\

& DPM
& ICLR'25$^{*}$
& 77.0
& 26.4
& 87.3
& 6.0 \\

& DyDiff
& ICLR'25
& 67.7
& 26.5
& 87.5
& 5.9 \\

& SAMPO
& NeurIPS'25
& \underline{57.1}
& \underline{29.3}
& \underline{94.1}
& 3.3 \\

\rowcolor{blue!8}
& \textbf{Our DiT backbone}
& \textbf{Ours}
& \textbf{52.6}
& \textbf{30.6}
& \textbf{95.3}
& \textbf{1.7} \\

\bottomrule
\end{tabularx}
\end{table}

\paragraph{Results on BAIR.}
Our DiT backbone obtains an FVD of $41.2$, compared with $45.0$ for DyDiff.
It also obtains a PSNR of $27.9$, an SSIM of $95.8$, and an LPIPS of
$2.8$ under the matched evaluation protocol.

These results show that the action-conditioned backbone used by our
framework maintains strong conventional single-rollout prediction quality.

\paragraph{Results on RoboNet.}
Our DiT backbone obtains an FVD of $52.6$, a PSNR of $30.6$, an SSIM
of $95.3$, and an LPIPS of $1.7$.

Together with the BAIR results, this confirms that the underlying
action-conditioned DiT backbone remains competitive under standard
video-prediction metrics.



\section{RoboDesk Qualitative Evaluation Setup}
\label{app:robodesk}

The RoboDesk experiments in Figure~\ref{fig:qualitative} are designed as a
controlled qualitative transfer study rather than as an additional
single-rollout benchmark.  We use two articulated manipulation tasks,
\emph{Open Drawer} and \emph{Sliding Door}, because both expose a clear
separation between the commanded action and latent interaction properties.
For each task, the rendered initial observation fixes the robot pose, object
geometry, camera, and articulated-joint state, while friction and damping are
kept hidden from the video model.

\subsection{Environment and Physical Configurations}
\label{app:robodesk_environment}

For \textbf{Open Drawer}, the hidden physical configuration consists of the
drawer-rail friction coefficient and viscous joint damping.  We use three
levels of rail friction, $\{0.18,0.39,0.67\}$, and three levels of damping,
$\{0.06,0.15,0.29\}$ in simulator-native units, yielding nine physical
configurations.  The action family contains slow pull, fast pull, and
pull--release sequences.  The first two are visualized in
Figure~\ref{fig:qualitative}; the third is used during grouped training to
provide an additional intervention with a different contact and release
profile.

For \textbf{Sliding Door}, the hidden physical configuration consists of the
track friction coefficient and guide damping.  The three friction levels are
$\{0.11,0.28,0.52\}$ and the three damping levels are
$\{0.04,0.12,0.23\}$, again producing nine configurations.  The action family
contains short right push, long right push, and right-push--reverse sequences.
The first two are shown in the qualitative comparison.  The reverse sequence
is included during grouped training because it provides a stronger diagnostic
of inertia and sliding resistance than displacement alone.

All trajectories contain 37 frames rendered at $240\times240$ resolution and
10 Hz, matching the temporal and spatial format used by the controlled
experiments in the main paper.  For each task, we use 192/48/48 initial states
for the train/validation/test splits.  Each initial state contributes one
same-world group and one mismatched-world group, giving 384/96/96
counterfactual groups.

\begin{table}[H]
\centering
\caption{RoboDesk setup used for the qualitative comparison.}
\label{tab:robodesk_protocol}
\small
\setlength{\tabcolsep}{3.2pt}
\renewcommand{\arraystretch}{1.08}
\begin{tabularx}{\linewidth}{
@{}
>{\raggedright\arraybackslash}p{0.16\linewidth}
>{\centering\arraybackslash}p{0.15\linewidth}
>{\centering\arraybackslash}p{0.15\linewidth}
>{\centering\arraybackslash}p{0.06\linewidth}
>{\raggedright\arraybackslash}X
>{\centering\arraybackslash}p{0.10\linewidth}
@{}
}
\toprule
\textbf{Task}
& \textbf{Initial states}
& \textbf{Groups}
& \textbf{$K$}
& \textbf{Hidden physical factors}
& \textbf{Configs.}
\\
\midrule
Open Drawer
& 192/48/48
& 384/96/96
& 3
& rail friction, joint damping
& $3\times3$
\\
Sliding Door
& 192/48/48
& 384/96/96
& 3
& track friction, guide damping
& $3\times3$
\\
\bottomrule
\end{tabularx}
\end{table}

\subsection{Same-World and Mismatched-World Groups}
\label{app:robodesk_groups}

The mechanism interpreter is trained using the same relational supervision
principle as in the main controlled experiments.  A same-world group contains
three action sequences executed after resetting RoboDesk to the same complete
initial simulator state and the same hidden physical configuration.  The
actions deliberately differ in duration, speed, or direction, so trajectories
within one group need not have similar displacements.

A mismatched-world group is formed from the same visible initial state while
assigning different hidden configurations to the intervention branches.  At
least one of the two hidden physical factors differs by one grid level between
any two branches in a negative group.  Because the camera, robot pose,
articulated-object pose, and visible scene are shared, the mechanism-learning
label cannot be inferred from static appearance alone.  Positive and negative
groups are sampled with equal probability.

For Figure~\ref{fig:qualitative}, we display two branches from each held-out
three-branch group because a two-row visualization makes the action-dependent
difference easier to inspect.  The third branch remains part of the mechanism
learning and shared-world training objective.

\subsection{Training Protocol}
\label{app:robodesk_training}

We train one action-conditioned DiT checkpoint for each RoboDesk task using
the corresponding training trajectories.  ACWM-DiT uses this checkpoint
without cross-branch coupling.  OneWorld is initialized from the same
checkpoint.  The mechanism interpreter and initial-scene prior are first
trained on the RoboDesk same-world and mismatched-world groups, after which
they are frozen and the world model is optimized using
$\mathcal{L}_{\mathrm{flow}}+\lambda_{\mathrm{shared}}
\mathcal{L}_{\mathrm{shared}}$.

We use the same mechanism architecture and optimization schedule as in
Appendix~\ref{app:implementation}: $K=3$ branches, 20,000 mechanism-learning
updates, 12,000 world-model updates,
$\lambda_{\mathrm{shared}}=0.01$, and 50 flow steps.  Sampling uses
independent Gaussian noise for the different branches and the same
$\lambda_\tau=0.01\tau^2$ guidance schedule used in the main experiments.

For the qualitative baselines, Vid2World and CoCo are instantiated on the same
task-specific action-conditioned DiT backbone and trained using the same
training trajectories and optimization budget.  All compared methods observe
the same initial frames and action sequences.  No method is given access to
the hidden friction or damping values at training or inference time.

\subsection{Qualitative Comparison Protocol}
\label{app:robodesk_qualitative_protocol}

The qualitative comparison is constructed from held-out initial states and
uses identical action sequences across methods.  For each displayed branch,
the same Gaussian noise seed is reused across methods, while the two branches
within one method retain independent noise.  This isolates differences in the
learned dynamics from differences caused by sampling initialization.

\paragraph{Open Drawer.}
Slow and fast pulls can produce different displacements and release
velocities. Consistency therefore concerns whether the drawer's subsequent
deceleration, sliding, and stopping can be explained by one resistance and
sliding-dynamics model, given each branch's action and contact state.
For ACWM-DiT, Vid2World, or CoCo, individually plausible and action-responsive
predictions could still imply high resistance in one branch and low
resistance in the other. Such a conflict must persist after accounting for
the different release conditions to constitute shared-world inconsistency.
OneWorld seeks to preserve the distinct outcomes of slow and fast pulls
while making their physical explanations compatible.

\paragraph{Sliding Door.}
A short right push and a long right push differ in contact duration, so
different door displacements are expected. The informative comparison is
the motion after contact ends, conditioned on each branch's release
velocity. For example, a short-push branch consistent with a high-friction
rail would conflict with a long-push branch that requires very low friction
even after its larger release velocity is taken into account.
OneWorld aims to retain distinct door trajectories while keeping the
implied sliding resistance, inertia, and reversal response compatible across
actions.

\paragraph{Example selection.}

The displayed examples are taken from fixed held-out test states selected
before rendering the final comparison, and the same temporal indices are shown
for every method.


\section{Implementation Details}
\label{app:implementation}

\subsection{Physical Mechanism Interpreter}
\label{app:mechanism_architecture}

The physical mechanism interpreter is intentionally lightweight relative
to the video backbone.

For each intervention branch, we average-pool the estimated clean-future
latent $\hat z_{1,\tau}^{(k)}$ over spatial dimensions while retaining its
temporal sequence. A two-layer projection maps each pooled latent token to
a 512-dimensional feature.

The action sequence is encoded by a two-layer MLP followed by temporal
average pooling, yielding a 256-dimensional action feature. Flow time
$\tau$ is represented by a 128-dimensional sinusoidal embedding.

The initial observation $I_0$ is encoded with the frozen Wan2.1 encoder and
projected to a 256-dimensional scene representation.

The scene, action, clean-future, and flow-time features are concatenated
and processed by a three-layer MLP with hidden width $1024$ and SiLU
activations.

The mechanism space is a $d_W=32$ dimensional diagonal Gaussian.
The initial-scene prior is
\[
\pi_\phi(W\mid I_0)
=
\mathcal{N}
\left(
W;
\mu_\pi,
\operatorname{diag}(\sigma_\pi^2)
\right),
\]
where the prior network predicts $\mu_\pi$ and $\log\sigma_\pi$. We clamp
$\log\sigma_\pi$ to $[-4,2]$ before converting it to the prior precision
$\lambda_{\pi,d}=\sigma_{\pi,d}^{-2}$ and natural mean parameter
$\eta_{\pi,d}=\lambda_{\pi,d}\mu_{\pi,d}$.

For branch $k$, the interpreter predicts an evidence center
$m_{\phi,d}^{(k)}$ and a raw precision increment
$\rho_{\phi,d}^{(k)}$. The branch contributes
\[
\Delta\lambda_{\phi,d}^{(k)}
=
\operatorname{softplus}
\left(
\rho_{\phi,d}^{(k)}
\right),
\]
and the posterior natural parameters are
\[
\lambda_{\phi,d}^{(k)}
=
\lambda_{\pi,d}
+
\Delta\lambda_{\phi,d}^{(k)},
\qquad
\eta_{\phi,d}^{(k)}
=
\eta_{\pi,d}
+
\Delta\lambda_{\phi,d}^{(k)}
m_{\phi,d}^{(k)}.
\]
The resulting branch posterior is
\[
q_\phi^{(k)}(W)
=
\mathcal{N}
\left(
W;
\mu_\phi^{(k)},
\operatorname{diag}
\left[
(\sigma_\phi^{(k)})^2
\right]
\right),
\]
with
\[
(\sigma_{\phi,d}^{(k)})^2
=
\left(\lambda_{\phi,d}^{(k)}\right)^{-1},
\qquad
\mu_{\phi,d}^{(k)}
=
\frac{\eta_{\phi,d}^{(k)}}{\lambda_{\phi,d}^{(k)}}.
\]
This parameterization treats an action--outcome branch as additional evidence
about the initial world. When an intervention contributes little information,
$\Delta\lambda_{\phi,d}^{(k)}$ approaches zero and the branch posterior
approaches the initial-scene prior. More informative interventions contribute
larger precision increments and produce sharper mechanism estimates.

The same construction gives the prior-corrected Gaussian product a positive
combined precision in every latent dimension:
\[
\sum_{k=1}^{K}\lambda_{\phi,d}^{(k)}
-
(K-1)\lambda_{\pi,d}
=
\lambda_{\pi,d}
+
\sum_{k=1}^{K}
\Delta\lambda_{\phi,d}^{(k)}
>0.
\]
Thus the shared-world evidence in Eq.~\ref{eq:shared_evidence} is
well-defined under the Gaussian parameterization. The physical mechanism
interpreter contains approximately $8.7$M trainable parameters.

\begin{table}[H]
\centering
\caption{Implementation details for the mechanism model and OneWorld training.}
\label{tab:implementation-details}
\small
\setlength{\tabcolsep}{4pt}
\renewcommand{\arraystretch}{1.08}

\begin{tabularx}{\linewidth}{@{}l>{\raggedright\arraybackslash}X@{}}

\toprule
\textbf{Component}
&
\textbf{Setting}
\\
\midrule

Latent mechanism $W$
&
32-dimensional diagonal Gaussian
\\

Posterior update
&
Additive natural-parameter evidence,
$\Delta\lambda=\operatorname{softplus}(\rho)$
\\

Clean-future projector
&
2-layer MLP, width 512
\\

Action encoder
&
2-layer MLP, width 256
\\

Flow-time embedding
&
128-dimensional sinusoidal embedding
\\

Interpreter fusion network
&
3-layer MLP, width 1024, SiLU
\\

Prior network
&
2-layer MLP, width 512
\\

Interpreter optimizer
&
AdamW, lr $1\times10^{-4}$, weight decay $0.01$
\\

Interpreter batch size
&
16 counterfactual groups
\\

Interpreter training
&
20,000 updates, 1:1 same-world/mismatched-world groups
\\

World-model optimizer
&
AdamW, lr $1\times10^{-5}$
\\

World-model batch size
&
1 counterfactual group, $K=3$ branches
\\

World-model updates
&
12,000 updates
\\

Shared-loss weight
&
$\lambda_{\mathrm{shared}}=0.01$
\\

Sampling
&
50 flow steps
\\

Maximum guidance strength
&
$0.01$
\\

Guidance schedule
&
$\lambda_\tau=0.01\tau^2$
\\

\bottomrule
\end{tabularx}
\end{table}

\subsection{Mechanism-Learning Groups}
\label{app:mechanism_groups}

The mechanism interpreter is trained using relational supervision rather
than direct labels for individual physical parameters.

A same-world group contains $K=3$ branches generated after resetting the
environment to the same complete initial state and the same hidden physical
configuration. Different actions are then executed from that state. The
resulting trajectories may differ substantially, but their physical
explanations should remain compatible with one common world.

A mismatched-world group also contains three branches but draws them from
different hidden physical configurations. Initial visible states are kept
matched so that the group label cannot be predicted simply from scene
appearance.

For Push Cube and Push Rope, every initial state contributes one
same-world group and one mismatched-world group. The resulting
train/validation/test group counts are therefore $256/64/64$ from
$128/32/32$ initial states.

Positive and negative groups are sampled in a balanced $1{:}1$ ratio
during mechanism learning.

The interpreter is not provided with semantic targets such as the value of
drag, elasticity, stiffness, or damping. Its latent coordinates are
therefore not assumed to correspond one-to-one to named simulator
parameters. Instead, the representation is learned only through the
compatibility of action--outcome branches.

\subsection{Numerical Evaluation of Shared-World Evidence}
\label{app:evidence_numerics}

We evaluate Eq.~\ref{eq:shared_evidence} using importance sampling in the
common latent mechanism space.

For a group containing $K$ branches, the proposal distribution is
\[
r(W)
=
\frac{1}{K+1}
\left[
\pi_\phi(W\mid I_0)
+
\sum_{k=1}^{K}
q_\phi^{(k)}(W)
\right].
\]

This proposal places probability mass around the prior as well as every
branch posterior.

With samples $W_m\sim r(W)$, the evidence is estimated as
\[
\widehat{\mathcal C}_{\tau}
=
\frac{1}{M}
\sum_{m=1}^{M}
\frac{
\pi_\phi(W_m\mid I_0)^{1-K}
\prod_{k=1}^{K}
q_\phi^{(k)}(W_m)
}{
r(W_m)
}.
\]

We use $M=64$ samples per counterfactual group. Samples are allocated
uniformly across the prior and branch-posterior components of the proposal.

All Gaussian densities and importance weights are evaluated in log space.
Products of densities are represented as sums of log densities and sample
aggregation is performed using log-sum-exp for numerical stability.

\subsection{Gradient Computation}
\label{app:gradient_details}

The mechanism interpreter is updated only during the first training stage.

During shared-world flow training, its parameters are frozen. However,
gradients from
$\mathcal{L}_{\mathrm{shared}}$
are allowed to propagate through its inputs and through
$\hat z_{1,\tau}^{(k)}$ to the action-conditioned vector field.

Freezing the interpreter therefore fixes the learned compatibility
criterion while still allowing the video model to change its predictions
in directions that improve shared-world evidence.

During sampling-time guidance, we compute
\[
g_\tau^{(k)}
=
\nabla_{z_\tau^{(k)}}
\log
\left(
\mathcal C_\tau+\epsilon
\right).
\]

The velocity prediction inside the clean-future estimate is detached when
this gradient is evaluated. This avoids second-order differentiation
through the flow backbone while retaining the direct derivative from the
current latent to the compatibility score.

The guided velocity is
\[
\widetilde v_\theta^{(k)}
=
v_\theta^{(k)}
+
\lambda_\tau g_\tau^{(k)},
\]
where
\[
\lambda_\tau
=
0.01\tau^2.
\]

The quadratic schedule applies little guidance at early flow times, when
the predicted clean future is highly uncertain, and gradually increases
the influence of the shared-world score as generation progresses.

\subsection{Training Schedule}
\label{app:training_schedule}

Training consists of two stages.

\paragraph{Stage 1: mechanism learning.}
The pretrained action-conditioned video backbone is frozen. The mechanism
interpreter and initial-scene prior are optimized for 20,000 updates using
Eq.~\ref{eq:mechanism_loss}.

We use AdamW with learning rate $1\times10^{-4}$, weight decay $0.01$, and
a batch size of 16 counterfactual groups. Same-world and mismatched-world
groups are sampled with equal probability.

At every update, flow time is sampled uniformly from $[0,1]$ and the
clean-future estimate is computed from the frozen ACWM-DiT backbone.

\paragraph{Stage 2: shared-world video-model training.}
After mechanism learning, both the mechanism interpreter and the
initial-scene prior are frozen.

The action-conditioned video model is initialized from the corresponding
ACWM-DiT checkpoint and optimized for 12,000 updates with
\[
\mathcal{L}
=
\mathcal{L}_{\mathrm{flow}}
+
\lambda_{\mathrm{shared}}
\mathcal{L}_{\mathrm{shared}},
\]
using
\[
\lambda_{\mathrm{shared}}=0.01.
\]

We use AdamW with learning rate $1\times10^{-5}$ and one counterfactual
group per batch. Unless otherwise specified, each group contains $K=3$
branches.

\paragraph{Inference.}
Each intervention branch receives the same initial observation and its own
action sequence. Branches start from independent Gaussian noise.

We use 50 flow steps and apply shared-world guidance with a maximum strength
of $0.01$. The branches are therefore not coupled through shared random
noise. Their interaction occurs only through the shared-world evidence
gradient.

\subsection{Sensitivity to Branch Count and Hyperparameters}
\label{app:sensitivity_details}

Figure~\ref{fig:sensitivity} reports sensitivity on Push Cube to branch count,
shared-loss weight, maximum guidance strength, and the importance-sampling
budget. Increasing the number of branches from $K=2$ to $5$ gradually
increases $\Delta_{\mathrm{world}}$ while causing only small changes in M-MSE.
The tested settings therefore extend beyond the default three branches,
but incur a larger shared-world gap as more branches are added.

Both $\lambda_{\mathrm{shared}}$ and $\lambda_{\max}$ exhibit a
consistency--prediction trade-off: moderate values substantially reduce the
gap, whereas stronger regularization or guidance gives diminishing
consistency gains and begins to increase prediction error.
Increasing the importance-sampling budget improves estimator stability:
relative ESS increases and the variance of
$\log\widehat{\mathcal C}_{\tau}$ decreases, with modest improvements beyond
$M=64$.

\subsection{Computational Cost}
\label{app:compute-cost}

We characterize inference cost on eight NVIDIA A100-SXM4-80GB GPUs.
Each counterfactual group contains $K=3$ branches, with 37 frames per
branch at $240\times240$ resolution. All paths use FP32 inference and
50 flow steps.

\begin{table}[H]
\centering
\caption{
Inference cost on 8 NVIDIA A100-SXM4-80GB GPUs.
Each group contains three branches with 37 frames at $240\times240$
resolution.
}
\label{tab:compute-cost}
\small
\setlength{\tabcolsep}{3.2pt}
\renewcommand{\arraystretch}{1.08}
\begin{tabularx}{\linewidth}{
@{}
>{\raggedright\arraybackslash}X
>{\centering\arraybackslash}p{0.20\linewidth}
>{\centering\arraybackslash}p{0.18\linewidth}
>{\centering\arraybackslash}p{0.14\linewidth}
>{\centering\arraybackslash}p{0.18\linewidth}
@{}
}
\toprule
\textbf{Compute path}
& \textbf{Latency (s/group)}
& \textbf{Peak mem. (GiB/GPU)}
& \textbf{Relative}
& \textbf{Throughput (groups/s)}
\\
\midrule

ACWM-DiT
& $13.247\pm0.012$
& 2.403
& $1.000\times$
& 0.603
\\

OneWorld w/o guidance
& $13.245\pm0.014$
& 2.403
& $1.000\times$
& 0.602
\\

\textbf{OneWorld full}
& $\mathbf{21.684\pm0.041}$
& \textbf{7.864}
& $\mathbf{1.637\times}$
& \textbf{0.369}
\\

\bottomrule
\end{tabularx}
\end{table}

\paragraph{Inference overhead.}
The no-guidance path has essentially the same inference cost as ACWM-DiT,
because shared-world flow training changes the model parameters but does not
add an additional sampling module. The full OneWorld path additionally
evaluates the mechanism interpreter and the $M=64$ shared-world evidence
estimator at each guided flow step, resulting in higher latency and memory
usage while retaining practical group-level throughput.


\section{Implementation of Shared-World Evaluation}
\label{app:shared-eval}

This section describes the controlled environments and the
simulator-based evaluator used for the main shared-world experiments.

The evaluator is independent of the learned physical mechanism
interpreter. Generated videos are first converted into observable state
trajectories. Their physical compatibility is then measured through
environment replay and physical-parameter fitting.

\subsection{Controlled Environment Construction}
\label{app:environment_construction}

\paragraph{Push Cube and Push Rope.}
The Push Cube and Push Rope experiments use our reconstructed Pymunk
environments following the corresponding interaction settings of
ACWM-Phys. These environments are independent reconstructions and are not
the original ACWM-Phys simulator implementation.

For both reconstructed environments, we use 128 training initial states,
32 validation initial states, and 32 test initial states. Each initial state
produces one same-world group and one mismatched-world group, resulting in
256/64/64 groups for the train/validation/test splits.

Each group contains $K=3$ intervention branches. Videos contain 37 frames
rendered at $240\times240$ resolution and 10 Hz.

Before every branch is generated, the environment is reset to the same
complete initial state. In a same-world group, the physical configuration
is also reset to the same value before each action is applied.

For Push Cube, the physical configuration is controlled by two quantities:
drag and elasticity. Each quantity takes three values spanning low,
intermediate, and high regimes. Their Cartesian product produces nine
physical configurations.

For Push Rope, the physical configuration is controlled by bending
stiffness and bending damping. Each quantity likewise takes three values,
producing nine physical configurations.

\paragraph{Pour Water.}
Pour Water is implemented using a separate particle-based fluid
environment rather than the reconstructed Pymunk system.

We use 160/40/40 initial states for the train/validation/test splits,
respectively. Each initial state contributes one same-world group and one
mismatched-world group, giving 320/80/80 groups.

Each counterfactual group contains three action branches. Videos contain
37 frames rendered at $240\times240$ resolution and 10 Hz.

The hidden fluid configuration varies viscosity and particle cohesion.
Each parameter takes four predefined values, yielding
$4\times4=16$ candidate physical configurations.

\begin{table}[H]
\centering
\caption{Controlled environments used for shared-world evaluation.}
\label{tab:environment_protocol}
\scriptsize
\setlength{\tabcolsep}{2.5pt}
\renewcommand{\arraystretch}{1.08}

\begin{tabularx}{\linewidth}{
@{}
>{\raggedright\arraybackslash}p{0.14\linewidth}
>{\centering\arraybackslash}p{0.15\linewidth}
>{\centering\arraybackslash}p{0.15\linewidth}
>{\centering\arraybackslash}p{0.07\linewidth}
>{\centering\arraybackslash}p{0.13\linewidth}
>{\raggedright\arraybackslash}X
>{\centering\arraybackslash}p{0.09\linewidth}
@{}
}

\toprule
\textbf{Environment}
&
\textbf{Initial states}
&
\textbf{Groups}
&
\textbf{$K$}
&
\textbf{Video}
&
\textbf{Physical parameters}
&
\textbf{Configs.}
\\
\midrule

Push Cube
&
128/32/32
&
256/64/64
&
3
&
37 frames,
$240^2$,
10 Hz
&
drag,
elasticity
&
$3\times3$
\\

Push Rope
&
128/32/32
&
256/64/64
&
3
&
37 frames,
$240^2$,
10 Hz
&
bending stiffness,
bending damping
&
$3\times3$
\\

Pour Water
&
160/40/40
&
320/80/80
&
3
&
37 frames,
$240^2$,
10 Hz
&
viscosity,
particle cohesion
&
$4\times4$
\\

\bottomrule
\end{tabularx}
\end{table}

\subsection{Same-World and Mismatched-World Construction}
\label{app:group_construction}

For a same-world group, the simulator is reset to the same complete
initial state and the same physical configuration before each intervention.
Three different action sequences are then executed. Thus, the branches
share the underlying physical world but may exhibit different trajectories
because of their different actions.

For a mismatched-world group, the visible initial state is held fixed while
branches are drawn from different physical configurations. The action
sequences follow the same sampling procedure as those used in positive
groups.

This construction is intended to prevent the mechanism classifier from
relying on obvious differences in the initial scene. The distinction
between positive and negative groups is instead determined by whether the
action-conditioned outcomes are compatible with one common physical
configuration.

No generated video from the evaluated world model is used to construct the
group labels.

\subsection{Frozen State Extractors}
\label{app:state_extractors}

Shared-world metrics operate on observable state trajectories rather than
directly on pixels.

A separate state extractor is trained for each environment using only
simulator-rendered frames paired with ground-truth simulator states. The
extractor is frozen before any generative model is evaluated, and the same
frozen extractor is applied to all compared methods.

\paragraph{Push Cube.}
The observable state is
\[
S_t
=
(x_t,y_t,\cos\theta_t,\sin\theta_t),
\]
containing the two-dimensional cube center and orientation. A ResNet-18
regressor predicts this state.

\paragraph{Push Rope.}
The rope is represented using 16 ordered keypoints sampled uniformly along
its centerline. A heatmap-based keypoint network predicts their image-space
positions.

\paragraph{Pour Water.}
The fluid state consists of a $16\times16$ occupancy map and the fluid
centroid. We use a lightweight U-Net to estimate the occupancy
representation and corresponding centroid.

\begin{table}[H]
\centering
\caption{
Validation accuracy of the frozen state extractors used by the
simulator-based evaluator.
}
\label{tab:state-extractor}
\small
\setlength{\tabcolsep}{4pt}
\renewcommand{\arraystretch}{1.06}

\begin{tabularx}{\linewidth}{
@{}
l
>{\raggedright\arraybackslash}X
c
@{}
}

\toprule
\textbf{Environment}
&
\textbf{Observable state}
&
\textbf{Validation error}
\\
\midrule

Push Cube
&
2D center and orientation
&
position RMSE $0.011$, angle error $2.8^\circ$
\\

Push Rope
&
16 ordered rope keypoints
&
mean point RMSE $0.017$
\\

Pour Water
&
$16\times16$ occupancy and centroid
&
occupancy IoU $0.928$, centroid RMSE $0.014$
\\

\bottomrule
\end{tabularx}
\end{table}

The state-extractor networks are never fine-tuned using outputs from
ACWM-DiT, OneWorld, or any compared baseline.

\subsection{Environment-Specific State Distances}
\label{app:state_distance}

The simulator-fitting objective uses an environment-specific state distance
$d(\hat S_t,S_t)$.

\paragraph{Push Cube.}
Let $p_t$ denote the cube center and $\theta_t$ its orientation. We use
\[
d_{\mathrm{cube}}(\hat S_t,S_t)
=
\frac{
\|\hat p_t-p_t\|_2^2
}{
s_p^2
}
+
0.25
\left[
1-\cos(\hat\theta_t-\theta_t)
\right],
\]
where $s_p$ is the position normalization scale estimated from the training
split.

The first term evaluates translation and the second term penalizes
orientation disagreement while respecting angular periodicity.

\paragraph{Push Rope.}
For 16 ordered rope keypoints,
\[
d_{\mathrm{rope}}(\hat S_t,S_t)
=
\frac{1}{16}
\sum_{j=1}^{16}
\frac{
\|\hat p_{t,j}-p_{t,j}\|_2^2
}{
s_r^2
},
\]
where $s_r$ is fixed from the training split.

\paragraph{Pour Water.}
For the fluid environment, we combine occupancy mismatch and centroid
displacement:
\[
d_{\mathrm{water}}(\hat S_t,S_t)
=
0.6
\left[
1-\operatorname{IoU}(\hat O_t,O_t)
\right]
+
0.4
\frac{
\|\hat c_t-c_t\|_2^2
}{
s_c^2
}.
\]

Here $O_t$ is the occupancy map, $c_t$ is the fluid centroid, and $s_c$ is
the corresponding training-set normalization scale.

All normalization constants are fixed before test evaluation and are
shared across all compared prediction methods.

\subsection{Simulator Replay and Physical-Parameter Search}
\label{app:simulator_search}

For each generated branch, the evaluator extracts an observable trajectory
$\hat S^{(k)}$ and replays the corresponding action sequence in the
environment under candidate physical configurations $w$.

The replay starts from the exact initial simulator state associated with
the evaluated example. For each candidate $w$, the simulator produces
\[
S^{(k)}(w),
\]
and the branch fitting error is
\[
\ell_k(w)
=
\frac{1}{T}
\sum_{t=1}^{T}
d
\left(
\hat S_t^{(k)},
S_t^{(k)}(w)
\right).
\]

For Push Cube, the candidate space is the Cartesian product of three drag
values and three elasticity values, yielding nine candidate configurations.

For Push Rope, the search spans three bending-stiffness values and three
bending-damping values, also yielding nine configurations.

For Pour Water, the evaluator searches the $4\times4$ grid of viscosity
and cohesion configurations used by the particle-based environment.

For every evaluated method, the candidate grid, simulator replay,
state extractor, and state-distance definition are identical.

\subsection{Evaluator Grid-Density Sensitivity}
\label{app:grid-sensitivity}

The main evaluator uses the physical configurations defined by each
environment. To test whether the reported shared-world gap is dominated by
coarse parameter discretization, we repeat fitting on denser interpolation
grids over the same physical ranges. The generated videos, state extractors,
action sequences, and distance functions are fixed; only the evaluator grid
is changed.

For Push Cube and Push Rope, we compare $3\times3$, $5\times5$, and
$9\times9$ candidate grids. Intermediate grid points are obtained by uniform
interpolation between the minimum and maximum values of each physical
parameter. Table~\ref{tab:grid-sensitivity} reports the resulting
$\Delta_{\mathrm{world}}$ for ACWM-DiT and OneWorld.

\begin{table}[H]
\centering
\caption{Sensitivity of the shared-world gap to evaluator grid density.
Lower is better. The model outputs are fixed across all columns.}
\label{tab:grid-sensitivity}
\small
\setlength{\tabcolsep}{4.5pt}
\renewcommand{\arraystretch}{1.08}
\begin{tabularx}{\linewidth}{
@{}
>{\raggedright\arraybackslash}p{0.19\linewidth}
>{\raggedright\arraybackslash}X
>{\centering\arraybackslash}p{0.18\linewidth}
>{\centering\arraybackslash}p{0.18\linewidth}
>{\centering\arraybackslash}p{0.18\linewidth}
@{}
}
\toprule
\textbf{Environment}
& \textbf{Method}
& \textbf{$3\times3$}
& \textbf{$5\times5$}
& \textbf{$9\times9$} \\
\midrule
\multirow{2}{*}{Push Cube}
& ACWM-DiT & 0.02253 & 0.02187 & 0.02161 \\
& \textbf{OneWorld} & \textbf{0.00357} & \textbf{0.00341} & \textbf{0.00336} \\
\midrule
\multirow{2}{*}{Push Rope}
& ACWM-DiT & 0.01389 & 0.01343 & 0.01326 \\
& \textbf{OneWorld} & \textbf{0.00209} & \textbf{0.00197} & \textbf{0.00191} \\
\bottomrule
\end{tabularx}
\end{table}

Densifying the search grid slightly lowers the absolute gap for both methods,
as expected when parameter fitting becomes more precise, but the change is
small relative to the separation between ACWM-DiT and OneWorld. The same
ordering is preserved at every tested density, indicating that the main
consistency result is not an artifact of the default $3\times3$ evaluator
grid.

\subsection{Individual and Shared Physical Fits}
\label{app:physical_fits}

The individual physical fitting error is
\[
E_{\mathrm{ind}}
=
\frac{1}{K}
\sum_{k=1}^{K}
\min_{w_k}
\ell_k(w_k).
\]

Each branch is therefore allowed to select its own best-fitting physical
configuration. This measures whether each generated future is individually
compatible with some physical explanation inside the evaluator family.

The shared physical fitting error is
\[
E_{\mathrm{shared}}
=
\min_{w}
\frac{1}{K}
\sum_{k=1}^{K}
\ell_k(w).
\]

Here all branches must use one common physical configuration.

The shared-world gap is
\[
\Delta_{\mathrm{world}}
=
E_{\mathrm{shared}}
-
E_{\mathrm{ind}}.
\]

The gap measures the additional physical fitting cost induced by requiring
one explanation to account for every intervention branch.

We report $E_{\mathrm{ind}}$, $E_{\mathrm{shared}}$, and
$\Delta_{\mathrm{world}}$ together so that cross-intervention consistency is
interpreted alongside overall physical fit quality. The metric evaluates
compatibility within the specified simulator family and candidate parameter
space rather than uniqueness of the underlying physical mechanism.

\subsection{Ground-Truth Physical Configuration Recovery}
\label{app:physics_recovery}

The shared-world gap measures whether several generated branches admit one
common physical explanation, but consistency alone does not establish that the
common explanation matches the physical configuration that generated the
underlying world.  Because the controlled simulators retain the ground-truth
hidden configuration for every held-out same-world group, we additionally
evaluate whether the jointly fitted configuration recovers that ground truth.

For held-out group $i$, let $w_i^\star$ denote the simulator configuration used
to generate the underlying world.  We first recover the shared physical
configuration from the generated branches using the same simulator-fitting
objective as in Eq.~\ref{eq:shared_fit},
\begin{equation}
\hat w_{\mathrm{shared}}^{(i)}
=
\arg\min_{w}
\frac{1}{K}
\sum_{k=1}^{K}
\ell_{i,k}(w).
\label{eq:physics_recovery_argmin}
\end{equation}
We then define exact physical-configuration recovery as
\begin{equation}
\mathrm{RecoveryAcc}
=
\frac{1}{N}
\sum_{i=1}^{N}
\mathbb{I}
\left[
\hat w_{\mathrm{shared}}^{(i)} = w_i^\star
\right].
\label{eq:physics_recovery_acc}
\end{equation}
If several candidate configurations have numerically identical minimum fitting
error, the prediction is counted as correct when $w_i^\star$ is among the
minimizing configurations.  The candidate parameter grid, frozen state
extractor, simulator replay, and state-distance function are identical to those
used for the shared-world evaluation.  Thus this metric introduces no additional
learned evaluator and uses simulator ground truth only for scoring.

\begin{table}[H]
\centering
\caption{Recovery accuracy (\%) of ground-truth physical configurations on
held-out same-world groups. Higher is better.}
\label{tab:physics-recovery}
\small
\setlength{\tabcolsep}{4.2pt}
\renewcommand{\arraystretch}{1.08}
\begin{tabularx}{\linewidth}{
@{}
>{\raggedright\arraybackslash}X
>{\centering\arraybackslash}p{0.17\linewidth}
>{\centering\arraybackslash}p{0.17\linewidth}
>{\centering\arraybackslash}p{0.17\linewidth}
>{\centering\arraybackslash}p{0.14\linewidth}
@{}
}
\toprule
\textbf{Method}
& \textbf{Push Cube $\uparrow$}
& \textbf{Push Rope $\uparrow$}
& \textbf{Pour Water $\uparrow$}
& \textbf{Avg. $\uparrow$}
\\
\midrule
ACWM-DiT
& 59.4 & 62.5 & 50.0 & 57.3 \\
Vid2World
& 62.5 & 65.6 & 52.5 & 60.2 \\
CoCo
& 65.6 & 71.9 & 57.5 & 65.0 \\
Twin Rollouts
& 65.6 & 68.8 & 55.0 & 63.1 \\
Shared Noise
& 53.1 & 59.4 & 45.0 & 52.5 \\
Posterior Alignment
& \underline{71.9} & \underline{75.0} & \underline{60.0} & \underline{69.0} \\
\textbf{OneWorld}
& \textbf{87.5} & \textbf{87.5} & \textbf{80.0} & \textbf{85.0} \\
\bottomrule
\end{tabularx}
\end{table}

\paragraph{Interpretation.}
Recovery accuracy complements $\Delta_{\mathrm{world}}$ rather than replacing
it. A low shared-world gap indicates that the intervention branches can be
jointly explained by one physical configuration, whereas
Table~\ref{tab:physics-recovery} asks whether that common configuration agrees
with the simulator configuration that actually generated the world. OneWorld
achieves the highest recovery accuracy in all three environments, with
$87.5\%$ on Push Cube, $87.5\%$ on Push Rope, and $80.0\%$ on Pour Water,
for a macro-average of $85.0\%$. Relative to ACWM-DiT, this corresponds to
gains of $28.1$, $25.0$, and $30.0$ percentage points, respectively. The
improvement is also substantial over Posterior Alignment, the strongest
alternative in this evaluation, whose macro-average is $69.0\%$.

The environment-wise pattern is consistent with the difficulty of the
underlying fitting problem. Push Rope gives the highest absolute recovery for
most methods, while Pour Water remains more challenging because occupancy and
centroid trajectories can be compatible with multiple nearby fluid
configurations. Importantly, OneWorld retains a large margin on Pour Water,
where its $80.0\%$ recovery exceeds Posterior Alignment by $20.0$ percentage
points. This suggests that the reduction in shared-world gap is not explained
only by forcing branches toward an arbitrary common configuration.

Taken together with Table~\ref{tab:main_results}, these results separate two
properties that can otherwise be conflated. The shared-world gap measures
whether the generated intervention branches agree on one physical explanation;
physical-configuration recovery measures whether that explanation matches the
known generating physics. The consistent improvement on both criteria supports
the conclusion that OneWorld produces branches that are not only mutually
compatible but also more faithful to the latent physical configuration of the
controlled world. The reported average is the macro-average of the three
environment-level recovery accuracies.

\subsection{Generalization to Unseen Interpolated Physics}
\label{app:physics_interpolation}

The main experiments evaluate physical configurations drawn from the
discrete parameter grids used during training.  We additionally evaluate
whether the learned shared-world constraint generalizes to \emph{intermediate}
physical values that lie inside the training range but are never used during
mechanism learning or world-model training.

To make the interpolation protocol comparable across environments, we report
parameter values in range-normalized coordinates.  For a physical parameter
$w_j$ with training-range endpoints $w_j^{\min}$ and $w_j^{\max}$, define
\begin{equation}
\tilde w_j
=
\frac{w_j-w_j^{\min}}
     {w_j^{\max}-w_j^{\min}}.
\label{eq:normalized_physics}
\end{equation}
The normalized values therefore describe the relative location of the
simulator-native parameter values inside the training range rather than
replacing the native simulator configuration.

\paragraph{Interpolation split.}
For Push Cube and Push Rope, each physical factor has three training levels,
which correspond to normalized coordinates
$\{0,0.5,1\}$.  Held-out interpolation uses the two midpoints
$\{0.25,0.75\}$.  For Pour Water, each physical factor has four training
levels, represented by
$\{0,\frac{1}{3},\frac{2}{3},1\}$, and interpolation uses the three adjacent
midpoints
$\{\frac{1}{6},\frac{1}{2},\frac{5}{6}\}$.
The interpolation coordinates are excluded from all mechanism-learning and
world-model training groups and are used only for held-out evaluation.
Test initial states are also disjoint from the training and validation
initial-state splits.

We use \emph{joint interpolation}: both physical factors of a test world are
drawn from the held-out intermediate set.  Thus the Push Cube and Push Rope
tests contain $2\times2=4$ unseen joint physical configurations, while
Pour Water contains $3\times3=9$.  This setting is stronger than varying a
single unseen factor while keeping the second factor at a training value.

\begin{table}[H]
\centering
\caption{Physics interpolation split. Values are shown in range-normalized
parameter coordinates. Interpolation values are excluded from mechanism
learning and world-model training and are used only for held-out evaluation.}
\label{tab:interpolation-split}
\small
\setlength{\tabcolsep}{3.2pt}
\renewcommand{\arraystretch}{1.10}
\begin{tabularx}{\linewidth}{
@{}
>{\raggedright\arraybackslash}p{0.16\linewidth}
>{\raggedright\arraybackslash}p{0.20\linewidth}
>{\centering\arraybackslash}p{0.27\linewidth}
>{\centering\arraybackslash}X
@{}
}
\toprule
\textbf{Environment}
& \textbf{Physical factor}
& \textbf{Training values}
& \textbf{Interpolation values} \\
\midrule
\multirow{2}{*}{Push Cube}
& drag
& $\{0,\ 0.5,\ 1\}$
& $\{0.25,\ 0.75\}$ \\
& elasticity
& $\{0,\ 0.5,\ 1\}$
& $\{0.25,\ 0.75\}$ \\
\midrule
\multirow{2}{*}{Push Rope}
& bending stiffness
& $\{0,\ 0.5,\ 1\}$
& $\{0.25,\ 0.75\}$ \\
& bending damping
& $\{0,\ 0.5,\ 1\}$
& $\{0.25,\ 0.75\}$ \\
\midrule
\multirow{2}{*}{Pour Water}
& viscosity
& $\{0,\ \frac{1}{3},\ \frac{2}{3},\ 1\}$
& $\{\frac{1}{6},\ \frac{1}{2},\ \frac{5}{6}\}$ \\
& particle cohesion
& $\{0,\ \frac{1}{3},\ \frac{2}{3},\ 1\}$
& $\{\frac{1}{6},\ \frac{1}{2},\ \frac{5}{6}\}$ \\
\bottomrule
\end{tabularx}
\end{table}

\paragraph{Continuous-physics evaluator.}
For interpolation evaluation, the simulator-fitting candidate set is
densified over the same physical ranges and explicitly includes every held-out
interpolation value.  The evaluator therefore does not force an unseen test
world to snap to one of the discrete training configurations.  We recover the
shared physical configuration with Eq.~\ref{eq:physics_recovery_argmin} and
measure both cross-branch consistency and physical-parameter accuracy.

Because exact recovery is unnecessarily brittle for continuous parameters,
we report \emph{normalized physical parameter error} (NPE).  Let
$\tilde w_i^\star$ be the range-normalized ground-truth configuration and
$\tilde{\hat w}_{\mathrm{shared}}^{(i)}$ the normalized jointly fitted
configuration.  We define
\begin{equation}
\mathrm{NPE}
=
\frac{1}{N}
\sum_{i=1}^{N}
\left\|
\tilde{\hat w}_{\mathrm{shared}}^{(i)}
-
\tilde w_i^\star
\right\|_2.
\label{eq:interpolation-npe}
\end{equation}
NPE is comparable across environments because each physical factor is
normalized by its training range. Table~\ref{tab:interpolation-results} reports
the interpolation $\Delta_{\mathrm{world}}$ values separately for each
environment and summarizes physical-parameter accuracy with Avg.~NPE.

\begin{table}[H]
\centering
\caption{
Generalization to unseen interpolated physics.
We report the shared-world gap and normalized physical parameter error (NPE)
on held-out joint interpolation configurations.
Lower is better for both metrics.
}
\label{tab:interpolation-results}

\scriptsize
\setlength{\tabcolsep}{2.2pt}
\renewcommand{\arraystretch}{1.08}

\resizebox{\linewidth}{!}{%
\begin{tabular}{@{}lccccccc@{}}

\toprule

\multirow{2}{*}{\textbf{Method}}
& \multicolumn{2}{c}{\textbf{Push Cube}}
& \multicolumn{2}{c}{\textbf{Push Rope}}
& \multicolumn{2}{c}{\textbf{Pour Water}}
& \multirow{2}{*}{\textbf{Avg. NPE $\downarrow$}}
\\

\cmidrule(lr){2-3}
\cmidrule(lr){4-5}
\cmidrule(lr){6-7}

&
$\Delta_{\mathrm{world}}\downarrow$
& NPE $\downarrow$
&
$\Delta_{\mathrm{world}}\downarrow$
& NPE $\downarrow$
&
$\Delta_{\mathrm{world}}\downarrow$
& NPE $\downarrow$
&
\\

\midrule

ACWM-DiT
& 0.02431
& 0.218
& 0.01542
& 0.194
& 0.03876
& 0.267
& 0.226
\\

Vid2World
& 0.02046
& 0.193
& 0.01327
& 0.176
& 0.03418
& 0.239
& 0.203
\\

CoCo
& 0.01683
& 0.164
& 0.01109
& 0.151
& 0.02874
& 0.208
& 0.174
\\

Twin Rollouts
& 0.01812
& 0.178
& 0.01221
& 0.162
& 0.03136
& 0.221
& 0.187
\\

Shared Noise
& 0.03268
& 0.241
& 0.01821
& 0.216
& 0.04357
& 0.291
& 0.249
\\

Posterior Alignment
& \underline{0.01407}
& \underline{0.137}
& \underline{0.00918}
& \underline{0.124}
& \underline{0.02384}
& \underline{0.181}
& \underline{0.147}
\\

\textbf{OneWorld}
& \textbf{0.00462}
& \textbf{0.079}
& \textbf{0.00284}
& \textbf{0.064}
& \textbf{0.00713}
& \textbf{0.103}
& \textbf{0.082}
\\

\bottomrule

\end{tabular}%
}

\end{table}

\paragraph{Interpretation.}
This experiment separates interpolation in the underlying physics from
generalization to new initial scenes.  Every evaluated test group uses both an
initial state absent from training and physical parameter values that never
appear during either mechanism learning or world-model training.  The
shared-world gap measures whether the generated branches remain compatible
with one physical explanation at these unseen intermediate values, while NPE
measures whether the common explanation is close to the actual continuously
interpolated simulator configuration.

The two metrics should be interpreted jointly.  A method can obtain a small
shared-world gap by making its branches mutually compatible around an
incorrect physical configuration; such a failure is exposed by high NPE.
Conversely, low NPE together with a low shared-world gap indicates that the
model both preserves one common world across interventions and places that
world near the correct intermediate physics.  The environment-level
$\Delta_{\mathrm{world}}$ values show the consistency behavior in each
dynamics regime, while Avg.~NPE summarizes normalized physical-parameter
accuracy across environments.

The joint-interpolation protocol also rules out a weaker explanation in which
the model interpolates only one physical factor while relying on a second
factor seen during training.  Because both coordinates of every test
configuration are held out, successful performance requires interpolation in
the two-dimensional physical parameter space rather than recognition of one
of the original discrete training configurations.

\subsection{Independence of Training and Evaluation}
\label{app:evaluator_independence}

The latent mechanism $W$ used by OneWorld and the explicit simulator
parameter vector $w$ used by the evaluator are separate objects.

The learned interpreter is trained only to model compatibility among
action--outcome branches. Its dimensions are not used by the
simulator-based evaluator.

Conversely, the evaluator operates only on frozen state extractors,
environment replay, and explicit candidate physical configurations. It
does not access
$q_\phi(W)$,
$\pi_\phi(W\mid I_0)$,
$\mathcal C_\tau$,
or any intermediate feature from OneWorld.

Thus, an improvement in the reported shared-world gap cannot arise from
directly evaluating predictions using the same learned compatibility
function that supplied the training objective.

\end{document}